\documentclass[10pt,conference]{IEEEtran}
\usepackage[T1]{fontenc}
\usepackage{microtype}
\usepackage{cite}
\usepackage{graphicx}
\usepackage{textcomp}
\usepackage[hyphens]{url}
\usepackage{booktabs}
\usepackage{array}
\usepackage{amsmath,amssymb,amsfonts,amsthm}
\usepackage{xcolor}
\usepackage{xspace}
\usepackage{tcolorbox}
\tcbuselibrary{breakable}
\usepackage{pifont}
\usepackage{multirow}
\usepackage{adjustbox}
\usepackage{tikz}
\usetikzlibrary{arrows.meta,positioning,fit,backgrounds,calc}
\usepackage{subfig}
\usepackage{hyperref}
\hypersetup{
    colorlinks=true,
    citecolor=blue,
    linkcolor=red,
    urlcolor=teal
}

\newcommand{\theHalgorithm}{\arabic{algorithm}}
\usepackage{algorithm}
\usepackage{algorithmic}

\newcommand{\sysname}{\textsc{VelServe}\xspace}
\newcommand{\cd}{c_{\mathrm{d}}}
\newcommand{\cp}{c_{\mathrm{p}}}
\newcommand{\SD}{S_{\mathrm{D}}}
\newcommand{\Nstar}{N^{\star}}
\newcommand{\tsolo}{t_{\mathrm{solo}}}
\renewcommand{\dbltopfraction}{0.8}
\newcounter{takeaway}
\newcommand{\takeaway}[2]{\refstepcounter{takeaway}\label{#1}%
\begin{tcolorbox}[colback=gray!5!white, colframe=gray!75!black,
  title=\textbf{Takeaway~\thetakeaway}, fonttitle=\small, fontupper=\small,
  before skip=4pt, after skip=3pt,
  left=4pt, right=4pt, top=2pt, bottom=2pt, toptitle=0.5pt, bottomtitle=0.5pt]
#2
\end{tcolorbox}}
\newtheorem{proposition}{Proposition}

\makeatletter
\newcommand{\tightdisplay}{%
  \setlength{\abovedisplayskip}{4pt plus 2pt minus 2pt}%
  \setlength{\belowdisplayskip}{4pt plus 2pt minus 2pt}%
  \setlength{\abovedisplayshortskip}{0pt plus 2pt}%
  \setlength{\belowdisplayshortskip}{3pt plus 2pt minus 1pt}}
\g@addto@macro\normalsize{\tightdisplay}
\AtBeginDocument{\tightdisplay}
\makeatother
\renewcommand{\dblfloatpagefraction}{0.8}
\ifdefined\pdfsuppressptexinfo\pdfsuppressptexinfo=-1\relax\fi

\hypersetup{pdftitle={VelServe: Taming Decode Jitters in LLM Inference with Tail-Calibrated Scheduling},
  pdfauthor={Ishaan Datta, Alind Khare, Koushik Kumar Dey, Ratnesh Jamidar, Anand Kumar Singh}}

\title{VelServe: Taming Decode Jitters in LLM Inference with Tail-Calibrated Scheduling}

\author{
  \IEEEauthorblockN{%
    \makebox[0.19\textwidth]{Ishaan Datta$^{\dagger}$}%
    \makebox[0.19\textwidth]{Alind Khare$^{\ddagger}$}%
    \makebox[0.19\textwidth]{Koushik Kumar Dey$^{\dagger}$}%
    \makebox[0.19\textwidth]{Ratnesh Jamidar$^{\dagger}$}%
    \makebox[0.19\textwidth]{Anand Kumar Singh$^{\dagger}$}}
  \IEEEauthorblockA{%
    \makebox[0.32\textwidth]{$^{\dagger}$Sprinklr}%
    \makebox[0.32\textwidth]{$^{\ddagger}$Microsoft}
    }
}

\begin{document}
\maketitle
\thispagestyle{empty}
\pagestyle{empty}
\makeatletter\global\@topnum\z@\makeatother

\begin{abstract}
Interactive LLM services must meet latency service-level objectives (SLOs) for the time to first token (TTFT) and the time between tokens (TBT), as streaming and soft real-time applications require predictable token delivery. We identify decode jitters in both aggregated and disaggregated serving. A jitter delays every request in the batch, and long responses span hundreds of steps, so even rare overruns cause many SLO violations. In aggregated serving, SLO-aware schedulers set each step's prefill budget until its predicted latency just meets the TBT target, leaving no room for mispredictions or host stalls. Disaggregation meets average TBT targets at a cost in capacity, yet the steps that admit new requests still run late.

We present \sysname, a scheduler that calibrates how much prefill work each decode step carries using the measured tail of its own iteration latency. The SLO determines how rare a late step must be, and \sysname leaves enough headroom in each iteration to prevent overruns and meet the target decode latency. This headroom slows prefill, so \sysname serves the shortest waiting prompt first. When TTFT deadlines grow with prompt length, the long prompts it delays have the most time to spare. Under disaggregation, it spaces admissions so that no request waits on more slow admission steps than its P99 target allows. On three attention families and varying workloads, \sysname sustains 2.4--2.9$\times$ the SLO goodput of the strongest baseline on Qwen3-8B and DeepSeek-V4-Flash, and at least 4.4$\times$ on DeepSeek-V2-Lite. With prefill and decode disaggregated on eight GPUs, \sysname meets the SLO for more than twice as many requests as state-of-the-art baselines.
\end{abstract}

\section{Introduction}
\label{sec:intro}

\begin{figure}[t]
\centering
\includegraphics[width=\columnwidth]{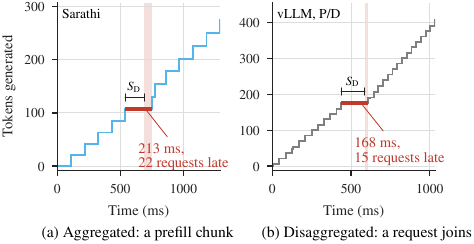}
\caption{Decode jitters persist despite chunked prefill and disaggregation (DeepSeek-V4-Flash, replayed Mooncake trace). Each step of a staircase is one iteration, which gives every decoding request one token. (a)~A Sarathi iteration that carries a prefill chunk runs past $\SD$, so every decoding request receives a late token at once. (b)~With prefill and decode disaggregated, the iteration that admits a newly prefilled request also runs past $\SD$. Shaded: time past $\SD$.}
\label{fig:intro-jitter}
\end{figure}

LLMs now power interactive applications such as chat and coding assistants, voice agents and multi-step agents. Agent histories and documents push prompts to $10^5$ tokens, and reasoning models produce long outputs, while short conversational requests remain most of the traffic. Providers serve these requests on shared GPUs, because a separate pool for each service would sit partly idle~\cite{cascade2026,goel2025niyama}. A serving system must therefore keep expensive GPUs busy while meeting each request's latency service-level objective (SLO).

An SLO bounds the time to first token (TTFT), set mainly by prefill, and the time between tokens (TBT), set by decode~\cite{agrawal2024sarathiserve,zhong2024distserve}. Providers size a deployment by its \emph{SLO goodput}: the highest request rate at which most requests still meet their SLO~\cite{zhong2024distserve}. This rate depends on the latency tail rather than the mean~\cite{li2026uniboost}. In streaming speech generation, for example, playback continues only if each audio chunk arrives before the previous one finishes playing~\cite{kamahori2026voxserve}. Such applications are judged by their slowest gaps, which a per-request mean such as the time per output token (TPOT) averages away~\cite{wang2024revisiting}. We therefore bound the 99th percentile of each request's own TBTs (\S\ref{sec:bg-slo}).

Prefill and decode interfere when they share a GPU, because a long prefill stalls every decoding request. Chunked prefill removes these generation stalls by mixing prompt chunks with decodes under a token budget~\cite{agrawal2024sarathiserve,kwon2023vllm}. Prefill/decode disaggregation runs the phases on separate GPUs~\cite{zhong2024distserve,patel2024splitwise,qin2024mooncake}, at a cost in capacity~\cite{gao2026duetserve}. In both designs, an iteration-level scheduler decides what each iteration contains.

The scheduler makes two decisions in each iteration. The \emph{order} decides which waiting requests are served first. Under first-come-first-served (FCFS) order, as in vLLM and Sarathi-Serve~\cite{kwon2023vllm,agrawal2024sarathiserve}, short requests wait behind long prefills, which is \emph{head-of-line blocking}. The \emph{token budget} caps the prefill work in each iteration. Every running request receives one token per iteration~\cite{yu2022orca}, so each iteration is a decode step for the whole batch. We call a decode step that runs longer than the TBT target a \emph{decode jitter}. A token budget bounds the work in each step, not its latency, so jitters persist under chunked prefill and disaggregation alike (Figure~\ref{fig:intro-jitter}). One jitter gives every decoding request a late token at once, and a request that decodes hundreds of tokens lives through hundreds of steps. As in the tail at scale~\cite{dean2013tail}, jitters that are rare per step therefore become common per request.

Decode jitters come from how prefill work is admitted into decode steps. SLO-aware schedulers avoid head-of-line blocking, but they fill each step with prefill until a latency model predicts that it will just meet the TBT target~\cite{agrawal2024medha,goel2025niyama,chen2025slosserve}. The step then jitters whenever the model underestimates its latency or the host stalls. Under disaggregation, the step that admits newly prefilled requests jitters in the deployment we measure (\S\ref{sec:motivation-pd}).

This paper asks how a scheduler can bound decode jitters for every request without giving up TTFT. We present \sysname, which bounds decode jitters in both colocated and disaggregated serving. It leaves each step a margin calibrated from the measured tail of its own iteration latency, and under disaggregation it limits how many admission steps each request waits on (\S\ref{sec:design}).

Across three attention families, \sysname sustains 2.4--2.9$\times$ the SLO goodput of the strongest baseline in aggregated serving under a 150-ms per-request P99 TBT target, and at least 4.4$\times$ on DeepSeek-V2-Lite. With prefill and decode disaggregated on eight GPUs, it meets the SLO for more than twice as many requests as vLLM and Niyama.

In summary, we make three contributions:
\begin{itemize}
\setlength\itemsep{0pt}
\item We identify decode jitters as the system-side cause of tail TBT in both aggregated and disaggregated serving. We trace them to how prefill is admitted into decode steps, and derive the jitter rate each TBT target tolerates (\S\ref{sec:motivation}).
\item We present \sysname, which turns the SLO into a tail-calibrated token budget and, under disaggregation, into an admission period (\S\ref{sec:design}). Added to Medha, its margin removes Medha's TBT misses.
\item We evaluate \sysname across three attention families, five workloads, six schedulers, TBT targets from 100 to 500\,ms, and a disaggregated deployment on eight GPUs (\S\ref{sec:eval}).
\end{itemize}

\section{Background and Motivation}
\label{sec:motivation}

\subsection{LLM Inference and Iteration-Level Scheduling}
\label{sec:bg-batching}
An LLM serves a request in two phases. \emph{Prefill} processes the whole prompt in one compute-bound pass and produces the first token. \emph{Decode} then produces one token per pass and is memory-bound, so engines batch the decodes of many requests together~\cite{agrawal2023sarathi,agrawal2024sarathiserve}. We call a decoding request a \emph{stream}. Modern engines schedule one \emph{iteration}, a forward pass over a changing batch, at a time~\cite{yu2022orca,kwon2023vllm}. Sarathi-Serve~\cite{agrawal2024sarathiserve} splits prompts into chunks, so a long prefill does not stall decoding. Each iteration then holds one decode token for every running request, plus prompt chunks up to a \emph{token budget} $N$, as in vLLM and SGLang~\cite{kwon2023vllm,zheng2024sglang}. \textbf{In every iteration, the scheduler makes two decisions. The budget size trades decode latency against prefill progress, while the ordering decides who waits.}

\subsection{Latency SLOs}
\label{sec:bg-slo}
The TTFT runs from a request's arrival to its first token. The TBT is the gap between two consecutive tokens. A request has one TTFT but hundreds of TBTs, so a TBT SLO must specify how to aggregate them. Most prior SLOs aggregate TBTs in ways that hide long tails. DistServe bounds each request's mean time per output token (TPOT)~\cite{zhong2024distserve}, and a mean spreads one long pause over hundreds of short gaps~\cite{wang2024revisiting}. Niyama gives every token a cumulative deadline~\cite{goel2025niyama}, which bounds a stream's average token rate and hides pauses in the same way. In our runs, such pauses last up to a second (Appendix~\ref{app:overrun}). Sarathi-Serve and Medha bound the 99th percentile over the tokens of all requests~\cite{agrawal2024sarathiserve,agrawal2024medha}. This pooled percentile lets many smooth streams hide the few that stall, just as one cluster-wide SLO number can hide the request classes that fail~\cite{cascade2026}. We therefore bound each request's own P99 TBT, which exceeds the target as soon as a request has more than a few long gaps.

\noindent\fbox{\parbox{\dimexpr\columnwidth-2\fboxsep-2\fboxrule}{\small
\textbf{SLO.} A request with prompt length $P$ \emph{meets the SLO} if its first token arrives within $\max(S_{F0},\,\kappa\,\tsolo(P))$ and its P99 TBT is at most $\SD$. Here $\tsolo(P)$ is the time to prefill $P$ tokens on an idle engine, and $\kappa$ is the \emph{TTFT SLO scale}. \emph{SLO attainment} is the fraction of requests that meet the SLO. \emph{SLO goodput} is the highest request rate at which attainment is at least 90\%.}}

\smallskip
The TTFT deadline scales with the prompt~\cite{chen2025slosserve,goel2025niyama}, because no fixed deadline fits both a short request and a prefill of many seconds. With a TTFT SLO scale of 3, a request may wait up to twice its own prefill time, a bound on its slowdown~\cite{bender1998stretch,harcholbalter2002}. The floor is $S_{F0}=1$\,s. We set $\SD=150$\,ms, faster than adults read~\cite{brysbaert2019reading}. A person reading buffered text may not notice a single late gap~\cite{liu2024andes,wang2024revisiting}, and the P99 target tolerates a few. A speech synthesizer or a downstream pipeline stage is delayed by every late gap, so we also report a stricter \emph{worst-gap} variant, in which every gap must be at most $\SD$.

\subsection{Scheduling Policies}
\label{sec:bg-policies}

\begin{table}[t]
\caption{The two decisions of each scheduler we study. Adaptive budgets add prefill until the predicted iteration latency reaches the limit shown.}
\label{tab:space}
\centering\footnotesize
\setlength{\tabcolsep}{4pt}
\begin{tabular}{@{}lll@{}}
\toprule
System & Token budget & Prompt order \\
\midrule
vLLM & fixed (8192 tokens) & FCFS \\
Sarathi & fixed, profiled for the SLO & FCFS \\
SPF & fixed & shortest prompt first \\
Medha & up to $\SD$, no margin & least slack \\
Niyama & up to slack, p99 margin$^{\dagger}$ & deadline + work left \\
\sysname & up to $\SD$ minus measured tail & shortest left$^{\ddagger}$ \\
\bottomrule
\end{tabular}
\par\smallskip{\scriptsize $^{\dagger}$Our addition (Appendix~\ref{app:niyama}). $^{\ddagger}$Likely misses go last.}
\end{table}

The order is a classical scheduling problem. First-come-first-served (FCFS) lets a long job block every short one behind it. Shortest-remaining-processing-time (SRPT) minimizes mean response time~\cite{schrage1968,smith1978}, and under overload, moving jobs that will miss anyway to the back protects the others~\cite{moore1968}. LLM schedulers pair one of these orders with a budget rule (Table~\ref{tab:space}). vLLM, Sarathi-Serve and SGLang's shortest-prefill-first (SPF)~\cite{sglang2026spf} fix the budget. The SLO-aware schedulers, Medha~\cite{agrawal2024medha} and Niyama~\cite{goel2025niyama}, fill each iteration up to a predicted latency, which describes the typical iteration better than the slow ones. Reordering cannot fix a slow iteration, because every stream in the batch waits for it.

\subsection{Iteration Cost}
\label{sec:motivation-regime}

\begin{figure}[t]
\centering
\includegraphics[width=\columnwidth]{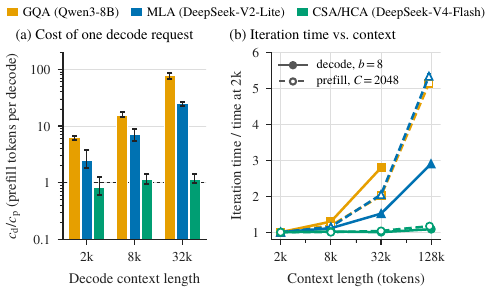}
\caption{With compressed attention, a decode costs about one prefill token (a) and context barely matters (b).}
\label{fig:cost}
\end{figure}

Both decisions need to know what work costs. The budget must predict how long an iteration will take, and the order must know how much work each request has left. We model an iteration with $b$ decodes and $p$ prefill tokens as
\begin{equation}
t = a + \cp\,p + \cd\,b + g\,\textstyle\sum_{i} C_i k_i + \epsilon,
\label{eq:cost}
\end{equation}
with a fixed cost $a$, per-token costs $\cp$ and $\cd$, a context term $g$ for each chunk of $C_i$ tokens at offset $k_i$, and a residual $\epsilon$ that the model does not explain, including host stalls. The context term is what separates attention architectures. Under GQA~\cite{ainslie2023gqa} and MLA~\cite{deepseek2024v2}, every token attends to its request's whole history. A chunk late in a long prompt, or a decode deep into a long answer, therefore costs far more than the same work early on (Figure~\ref{fig:cost}b). Compressed attention, as in DeepSeek-V4~\cite{deepseek2026v4}, attends to a bounded set of pooled entries instead. Its cost barely grows with context, and a decode costs about as much as one prefill token (Figure~\ref{fig:cost}a).

\takeaway{tk:cost}{On compressed attention, token counts predict iteration latency well, because cost barely grows with context. Under GQA and MLA, a cost estimate must also account for each token's context.}

\subsection{Failure Mode 1: Head-of-Line Blocking}
\label{sec:motivation-hol}

FCFS engines face a dilemma when long and short prompts share a GPU. With a large budget, an iteration that carries a chunk of a long prompt lasts close to a second, and every stream in the batch waits for it. A static budget sized for the TBT target, as in Sarathi-Serve, removes these long iterations. But each long prefill now spans many more iterations, and a short request that arrives behind it still waits for all of them. Short requests then spend most of their TTFT waiting behind other prefills (Appendix Figure~\ref{fig:hol}). A smaller budget only moves the delay from decode tokens to first tokens.

\takeaway{tk:order}{Tuning the token budget cannot remove head-of-line blocking. Removing it requires serving short prompts ahead of long ones.}

\subsection{Failure Mode 2: Decode Jitters}
\label{sec:motivation-tail}

\begin{figure}[t]
\centering
\includegraphics[width=\columnwidth]{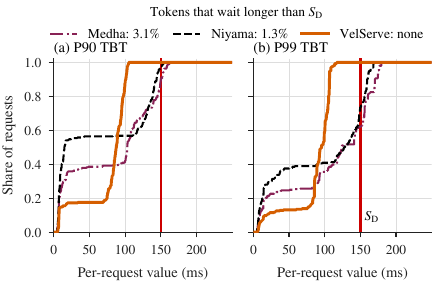}
\caption{Decode jitters stay out of each request's P90 TBT (a) but reach its P99 (b), so far fewer requests meet $\SD$.}
\label{fig:fanout}

\end{figure}

Each iteration is a decode step for every running stream, so its latency is a TBT for each of them. We call an iteration that runs longer than $\SD$ a \emph{decode jitter}. Jitters come from the interplay between prefill and decode. A decode step that carries prefill runs long when the scheduler admits more prefill than the step can absorb, whether by plan, by misprediction or because the host stalls. One jitter is amplified in two directions. Across the batch, it gives every stream a late token at once. Across time, a stream waits on every decode step it lives through, often hundreds. This is the tail at scale~\cite{dean2013tail}: delays that are rare in each component become common for a request that depends on many components. Even if jitters were independent and occurred in 1\% of steps, 95\% of 300-token answers would meet at least one (Appendix~\ref{app:analysis}). Real jitters are also correlated, because a host stall delays a whole batch and stalls arrive in bursts. One burst can give a request more late gaps than its P99 target allows. Figure~\ref{fig:fanout} shows this amplification under two SLO-aware schedulers (DeepSeek-V2-Lite, Mixed, 1.25\,req/s), plotting each request's P90 and P99 TBT. Only a few percent of their tokens wait on a jitter, and \sysname keeps every request within $\SD$. At a 150-ms bound, 89--98\% of requests meet the bound at P90, but only 62--74\% meet it at P99. The bound is the same in both panels. Jitters are rare enough to stay out of each request's P90 but common enough to reach its P99.

\begin{table}[t]
\caption{Iterations (\%) planned / measured to run past $\SD$ (Mooncake-shaped lengths). Medha plans none, yet still jitters.}
\label{tab:overrun}
\centering\footnotesize
\setlength{\tabcolsep}{5pt}
\begin{tabular}{@{}lccc@{}}
\toprule
System & Qwen3-8B & DS-V2-Lite & DS-V4-Flash\\
\midrule
Niyama & 1.71 / 2.15 & 1.00 / 1.11 & 8.37 / 12.40\\
Medha & 0.00 / 1.36 & 0.00 / 1.60 & 0.00 / 3.97\\
\sysname & 0.00 / 0.00 & 0.00 / 0.01 & 0.00 / 0.02\\
\bottomrule
\end{tabular}
\end{table}

\textbf{Steps filled to a predicted latency often jitter.} A scheduler that fills each decode step until its predicted latency reaches $\SD$ leaves no time for errors. A \emph{margin}, time left unused in each planned step, can absorb them. Medha keeps no margin, and 1.4--4\% of its iterations jitter, more than a P99 target tolerates (Table~\ref{tab:overrun}). A margin helps only if it matches the iterations being run. The prediction error depends on what each iteration contains, and the budget decides that. Our Niyama implementation measures its margin on static-budget runs, so its margin describes the wrong iterations. Cumulative token deadlines cause a separate problem, because they bound a stream's average token rate rather than each gap. A stream that runs ahead banks slack, and Niyama spends it on decode steps longer than $\SD$, some lasting a second (Appendix~\ref{app:overrun}). Niyama plans 1--8\% of its iterations as jitters (Table~\ref{tab:overrun}). No error margin prevents these jitters, because they are planned.

\takeaway{tk:budget}{Schedulers that fill each decode step to its predicted latency cause decode jitters. Each jitter delays every stream in the batch at once, so even rare jitters break strict TBT targets.}

\subsection{Decode Jitters under Disaggregation}
\label{sec:motivation-pd}

Disaggregation removes prompt chunks from decode steps~\cite{zhong2024distserve,patel2024splitwise}, so disaggregated systems meet average TPOT targets. The interplay between prefill and decode remains, because each prefilled request must still enter the decode instance. We call the decode step that admits newly prefilled requests a \emph{join}. In the expert-parallel deployment we study (\S\ref{sec:eval-disagg}), a join takes 170--185\,ms however many requests it admits, so every join is a decode jitter. Expert-parallel ranks advance in lockstep~\cite{vllm2026dp}, so a join on any rank delays the next token of every stream on every rank. A stream that decodes for many seconds also waits through many joins. Under vLLM, every request meets a 150-ms mean TPOT target, yet most requests miss the P99 TBT target (Figure~\ref{fig:pd}a). A mean TPOT target cannot see the joins, but a per-request P99 target counts them.

\takeaway{tk:pd}{In our disaggregated deployment, the decode step that admits new requests is a decode jitter that no token budget can shorten. The scheduler can instead control how often each request experiences such a jitter.}

\section{\sysname}
\label{sec:design}

\subsection{Overview}
\label{sec:design-overview}

\begin{figure}[t]
\centering
\includegraphics[width=\columnwidth]{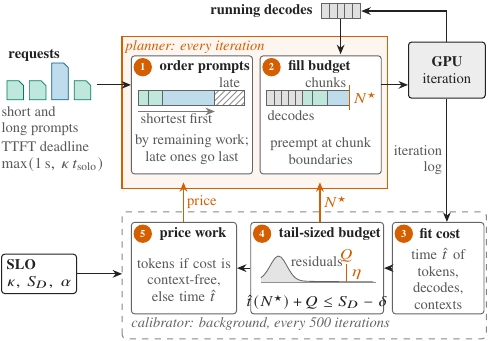}
\caption{Overview of \sysname. The planner \ding{172}--\ding{173} runs every iteration, and the calibrator \ding{174}--\ding{176} runs in the background.}
\label{fig:design}
\end{figure}

\sysname is a vLLM scheduler, configured by the SLO and an attainment target $\alpha$, that addresses Takeaways~\ref{tk:cost}--\ref{tk:budget}. It has a fast path, run every iteration, and a slow path, run in the background (Figure~\ref{fig:design}). In every iteration, the planner \ding{172}~orders waiting prompts, shortest first (\S\ref{sec:design-order}), and \ding{173}~fills the token budget $\Nstar$ with one token per running request, plus prefill chunks in that order. vLLM runs the batch and logs its time and contents. Every 500 iterations, the calibrator \ding{174}~refits the cost model of Eq.~\eqref{eq:cost} to that log. It \ding{175}~sizes $\Nstar$ with a margin for the tail (\S\ref{sec:design-budget}) and \ding{176}~decides whether to estimate cost from token counts or from a context-aware time model (\S\ref{sec:design-cost}). Under disaggregation, the same principle decides when the decode instance admits newly prefilled requests (Takeaway~\ref{tk:pd}, \S\ref{sec:design-disagg}).

\subsection{An Illustrative Example}
\label{sec:design-example}

\begin{figure*}[t]
\centering
\includegraphics[width=0.85\textwidth]{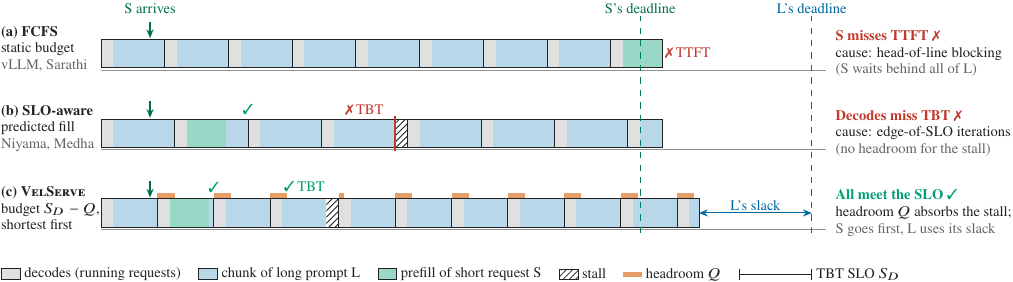}
\caption{How a short request S and a host stall fare under FCFS, SLO-aware scheduling and \sysname (schematic, one box per iteration).}
\label{fig:timeline}
\end{figure*}

In Figure~\ref{fig:timeline}, a short request S arrives behind a long prompt L that is partway through its prefill, and all three schedulers face one host stall. \textbf{(a)}~FCFS keeps every iteration within its static budget, but S waits behind L's remaining chunks and misses its TTFT deadline. \textbf{(b)}~SLO-aware schedulers start S at once, but they fill each iteration to its predicted latency, so the iteration that meets the stall becomes a decode jitter. \textbf{(c)}~\sysname plans each iteration to $\SD-Q$, where $Q$ is its margin, so the stall fits within the margin. The margin slows prefill, so \sysname spends L's slack: S preempts L at a chunk boundary, and L still meets its later, prompt-scaled deadline.

\vspace{-4pt}\subsection{Tail-Sized Budget}
\label{sec:design-budget}

\textbf{Problem.} A request decodes hundreds of tokens, and each decode step it lives through can be a decode jitter (\S\ref{sec:motivation-tail}). The budget must therefore guard against the far tail of iteration times. The tail comes from mispredicted prefill-heavy iterations and from host stalls.

\textbf{Key insight.} The SLO determines how rare an overrun must be. The scheduler's own iterations determine how large a margin keeps overruns that rare, because the budget decides what those iterations contain. \sysname therefore sizes the margin from running statistics of its own residuals, as recent work does for request order~\cite{li2026uniboost}.

\textbf{Mechanism.} \sysname treats both sources as the residual $\epsilon$ of Eq.~\eqref{eq:cost}. Let $\hat t(N)$ be the predicted time of a full iteration with budget $N$ at a heavy reference load. Let $Q$ be a margin that the residual exceeds in a share $\eta$ of iterations. \sysname chooses
\begin{equation}
\begin{gathered}
\Nstar=\max\{N:\ \hat t(N)+Q\le \SD-\delta\},\\
\eta=\beta(1-\alpha)(k+1)/\bar o,
\end{gathered}
\label{eq:tail}
\end{equation}
where $\delta$ covers token delivery, $\bar o$ is the mean output length, $k$ is the number of late gaps a request may have and still meet its TBT target, and $\beta=0.5$ is the share of the miss budget given to TBT. The budget is thus the largest one whose predicted latency plus the margin fits the TBT target. With overrun rate $\eta$, a request has $\bar o\eta$ late gaps on average. By Markov's inequality, it has more than $k$ of them with probability at most $\bar o\eta/(k+1)$, which we set to $\beta(1-\alpha)$ (Appendix~\ref{app:analysis}). The bound needs no independence between iterations, but it requires that the iterations a request experiences overrun at rate at most $\eta$, within the reference load. Every colocated run uses $k=0$, the worst-gap depth, so one margin covers both SLOs. The attainment target $\alpha=0.99$ is a calibration target, distinct from the 90\% that defines SLO goodput.

\begin{figure}[t]
\centering
\includegraphics[width=\columnwidth]{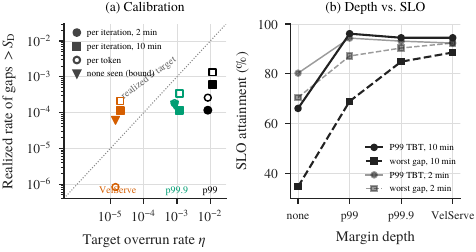}
\caption{Margin rules on DeepSeek-V4-Flash. (a) Each rule overruns less often than its target (triangles: no overrun observed). (b) A p99 margin meets the P99 SLO, but the worst-gap bound needs \sysname's deeper margin.}
\label{fig:tail}
\end{figure}

Figure~\ref{fig:tail} shows what the margin's \emph{depth}, the quantile of the residuals at which it is set, contributes. Every rule overruns less often than its target (a). Under the P99 target, a measured margin suffices in the tested setting, whereas no margin, as in Medha, does not (b). Under the worst-gap bound, greater depth decides compliance.

\begin{algorithm}[t]
\caption{Margin and budget calibration, run in the background every 500 iterations (full version: Appendix Algorithm~\ref{alg:calib}).}
\label{alg:calib-main}
\small\raggedright
\begin{algorithmic}[1]
\REQUIRE iteration log; SLO $\SD$; attainment target $\alpha$; late gaps allowed per request $k$
\STATE fit the cost model $\hat t$ of Eq.~\eqref{eq:cost} to recent iterations
\STATE $\epsilon_s\gets t_s-\hat t(s)$ for each recent iteration $s$ \COMMENT{residuals}
\STATE $\eta\gets\beta(1-\alpha)(k+1)/\bar o$ \COMMENT{Eq.~\eqref{eq:tail}}
\STATE $f_{\min}\gets$ share of residuals that would break $\SD$ even at the smallest budget
\STATE $Q\gets$ the residual exceeded by a share $\eta+f_{\min}$ of recent iterations
\STATE $\Nstar\gets\max\{N:\hat t(N)+Q\le\SD-\delta\}$ at the reference load
\STATE \textbf{if} $\Nstar$ reaches its minimum, flag the TBT SLO as infeasible
\end{algorithmic}
\end{algorithm}

\textbf{Estimating the margin.} Calibration forms a feedback loop. The margin sets the budget, the budget sets what each iteration contains, and those contents set the residuals from which the next margin is computed. Algorithm~\ref{alg:calib-main} therefore computes $Q$ from \sysname's own recent iterations. $Q$ is a rank statistic (line 5), because the tail is a light bulk plus rare clusters of host stalls that a fitted distribution misses. Stalls that would break $\SD$ even at the smallest budget count toward the rank (line 4) rather than raising $Q$, because no budget can absorb them. On Qwen3-8B, \sysname's own iterations under load raise the margin from its starting value of 2.3\,ms to about 50\,ms (\S\ref{sec:eval-ablation}).

\subsection{Shortest-First Order}
\label{sec:design-order}

\textbf{Problem.} The margin slows prefill, and therefore raises TTFT.

\textbf{Key insight.} Under a length-normalized deadline, a request's slack grows with its own prefill time. A short request waiting behind a long one can lose its whole deadline, while a long prompt delayed by a short one loses a small fraction of its own.

\textbf{Mechanism.} After the decodes, the planner fills the budget with chunks of the prompts that have the fewest remaining tokens (or the least predicted remaining time, \S\ref{sec:design-cost}). A shorter arrival preempts a running prefill at a chunk boundary. Its KV cache stays on the GPU, so preemption only adds delay. Chunked prefill thus behaves like a preemptive single server. For such a server, serving the shortest remaining job first minimizes the sum of response times~\cite{schrage1968,smith1978}. The result carries over to the sum of TTFTs if iteration cost is linear in tokens and independent of context, and if decode batch sizes are held fixed (Appendix~\ref{app:analysis}, Proposition~\ref{prop:srpt}). In a real schedule, the order also changes when prefills start decoding, so we test the full system by experiment. When even an idle engine could no longer meet a request's deadline, \sysname moves the request behind the others. It is still served, as in the Moore--Hodgson rule~\cite{moore1968}. Neither policy is new~\cite{wu2023fastserve,goel2025niyama}.

\textbf{Prompt-scaled deadlines limit the cost of shortest-first order.} Shortest-first order can starve the largest jobs~\cite{li2026uniboost}, and size-based LLM schedulers fail long requests far more often than short ones~\cite{cascade2026}. Two properties limit this cost in \sysname. First, prompt lengths are known at arrival, so the order needs no length prediction. Second, the TTFT deadline grows with the prompt, so a long prompt's slack grows with the work that can overtake it. Under SRPT, a long prompt still meets its deadline while shorter work uses at most about $1-1/\kappa$ of the prefill capacity, two thirds at $\kappa=3$ (Appendix~\ref{app:analysis}). Above that load, the deadline-aware baselines also time out on long prompts (\S\ref{sec:eval-sd}).

\subsection{Cost from Token Counts}
\label{sec:design-cost}

\textbf{Problem.} Both the budget and the order need the cost of an iteration and of a request's remaining work. SLO-aware schedulers estimate it with context-aware latency predictors~\cite{agrawal2024medha,goel2025niyama}.

\textbf{Key insight.} Under compressed attention, iteration latency is nearly linear in token counts and independent of context (\S\ref{sec:motivation-regime}). An iteration with $b$ decodes can then carry $\Nstar-w\,b$ prefill tokens, with a fixed decode weight $w=\cd/\cp$, and remaining prompt tokens order requests nearly as well as remaining time (Appendix~\ref{app:analysis}). On DeepSeek-V4-Flash, token counts track cost up to about 170--220k tokens of context.

\textbf{Mechanism.} \sysname fits Eq.~\eqref{eq:cost} online by a Huber regression on the last 8k iterations, regularized toward an offline fit, and tests the two conditions above: whether costs are context-free, and whether a decode weighs about one prefill token. If both hold, it uses a plain token budget and sizes requests in tokens. Otherwise it budgets each iteration and sizes each request with the fitted time model. On DeepSeek-V4-Flash, costs are context-free but a decode weighs more than one prefill token, so the model reduces to a weighted token budget. On DeepSeek-V2-Lite, token counts would misjudge iterations deep in long contexts and lower attainment at 0.75\,req/s from 99\% to 84\% (Appendix~\ref{app:space}). With prefix caching, a follow-up turn recomputes only its new tokens, so \sysname sizes each waiting request by its uncached suffix (Appendix~\ref{app:design}). On multi-turn sessions, this keeps follow-up turns within a P99 TTFT of 1.5\,s up to saturation (Appendix~\ref{app:multiturn}).

\textbf{Implementation.} \sysname is a plugin for vLLM~0.30's V1 scheduler of about 1k lines of Python. It reorders vLLM's queues and caps each request's chunk, so batching, KV allocation, CUDA graphs and asynchronous scheduling are unchanged. Planning takes 0.01\,ms per iteration at the median, against a median iteration time of 10.6\,ms on DeepSeek-V4-Flash.

\subsection{Disaggregated Serving}
\label{sec:design-disagg}

\textbf{Problem.} Under disaggregation, joins cause the decode jitters (\S\ref{sec:motivation-pd}), so a request meets its P99 TBT target only if it waits through few of them.

\textbf{Key insight.} A join cannot be made shorter, but the scheduler decides when joins happen. The P99 target tolerates a few late gaps per request (Eq.~\eqref{eq:tail}). A request with $n$ output tokens may have $k(n)=(n-2)-\lfloor0.99(n-2)\rfloor$ gaps above $\SD$, which is six for a 600-token answer. At decode step time $t_d$, the stream decodes for $n\,t_d$ seconds, so it can absorb one stall every $n\,t_d/k(n)$ seconds. In colocated serving, \sysname limits how long each iteration runs (\S\ref{sec:design-budget}). Under disaggregation, it instead limits how often each stream is stalled by a join. Two classical ideas keep streams within this rate. First, a join costs the same however many requests it admits, so admitting waiting requests together amortizes it, as group commit amortizes one log write over many transactions~\cite{dewitt1984}. Second, ranks in lockstep share every stall. Admitting on all ranks at once therefore merges their joins into one stall, as coordinated garbage collection merges pauses across nodes~\cite{maas2015trashday}. vLLM already aligns prefill admission across expert-parallel ranks at a fixed interval~\cite{vllm2026cadence}, and \sysname derives this interval from the SLO.

\textbf{Mechanism.} Each decode rank holds requests whose KV cache has arrived until the next \emph{epoch}, a wall-clock boundary shared by all ranks. The epoch period $T$ is the median of $n\,t_d/k(n)$ over recently completed requests, so no output length is predicted. It is rounded up on a power-of-two grid, which keeps ranks aligned even when their estimates differ slightly. If a join fits within $\SD$, the gate stays open, and \sysname admits requests as vLLM does.

\textbf{TTFT deadlines.} The first token reaches the client from the decode instance, so holding a request spends its TTFT slack, as the margin does in colocated serving. As there (\S\ref{sec:design-order}), the order absorbs this cost. The prefill instance keeps shortest-first order and forwards each request's TTFT deadline with its KV cache. The decode stage then schedules against the request's end-to-end deadline, as each stage of a distributed real-time task does~\cite{kao1997deadline}. A request that cannot wait for the next epoch opens an earlier window, and ready requests on other ranks join the same stall (Appendix~\ref{app:disagg}). These windows spend part of the TBT allowance on first tokens, so the bound of Eq.~\eqref{eq:tail} does not carry over.

\begin{figure*}[p]
\centering
\includegraphics[width=\textwidth]{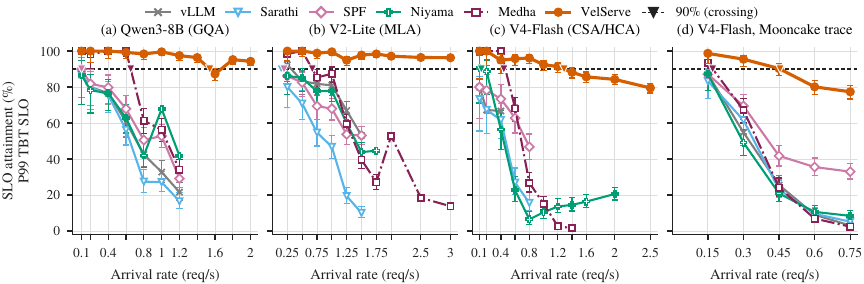}
\caption{SLO attainment vs.\ load (TTFT deadline and per-request P99 TBT $\le 150$\,ms). (a--c) Mixed; (d) replayed Mooncake trace, four slices pooled. Error bars: 95\% Wilson interval over pooled requests.}
\label{fig:main}
\end{figure*}

\begin{figure*}[p]
\centering
\includegraphics[width=\textwidth]{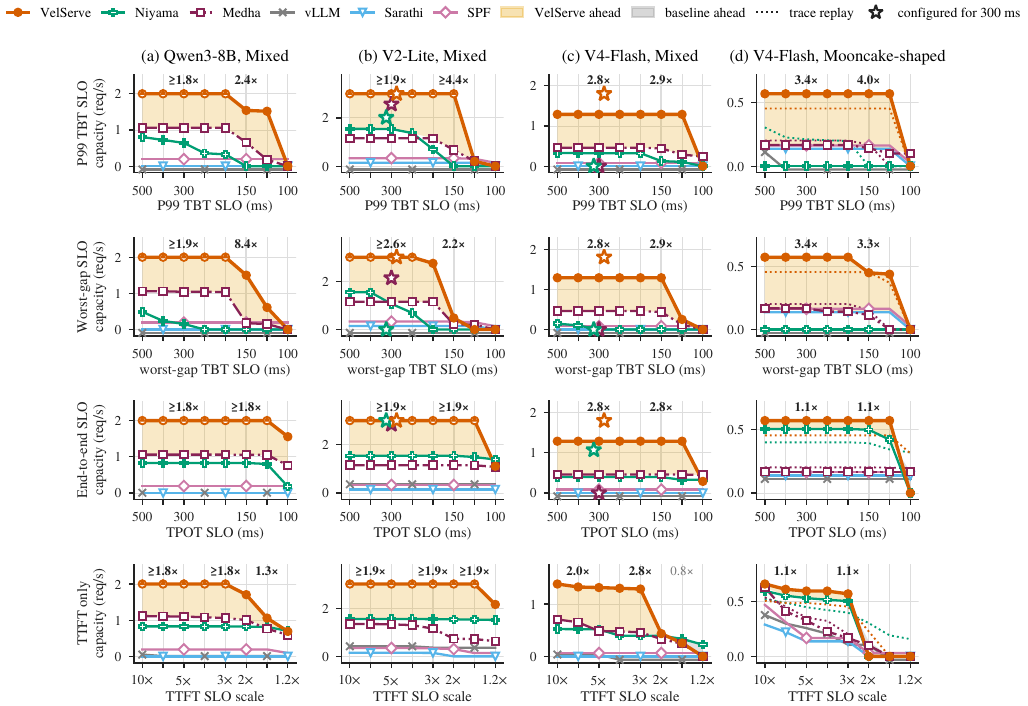}
\caption{SLO goodput when each part of the SLO is tightened or relaxed, re-scoring the same runs (configured for 150\,ms and a TTFT SLO scale of 3). Rows: (1)~per-request P99 TBT bound; (2)~worst-gap bound; (3)~end-to-end deadline, with the TPOT SLO swept; (4)~TTFT only, with the TTFT SLO scale $\kappa$ swept. \sysname leads under both TBT bounds down to 150\,ms, and rows 3--4 show that its margin costs no end-to-end or first-token goodput. Labels give \sysname's SLO goodput over the best baseline's, and stars mark runs configured for 300\,ms. In (d), the trace is dotted, and Niyama ran from 0.4\,req/s.}
\label{fig:sdgap}
\end{figure*}

\section{Evaluation}
\label{sec:eval}

\subsection{Setup}
\label{sec:eval-setup}

\textbf{Scope.} We evaluate two deployments, and every system runs in the same vLLM build. Aggregated serving runs each model on an H100 GPU, and disaggregated serving uses a node of eight RTX PRO 6000 GPUs (\S\ref{sec:eval-disagg}). Three models cover the three attention families of \S\ref{sec:motivation-regime}: Qwen3-8B (GQA), DeepSeek-V2-Lite (MLA) and DeepSeek-V4-Flash (compressed attention). Aggregated DeepSeek-V4-Flash runs use one 4-way expert-parallel shard, whose per-token compute and weight traffic match a shard of a full deployment (Appendix~\ref{app:setup}). The main workload, \emph{Mixed}, has Poisson arrivals. Of its prompts, 75\% have 128--2k tokens, 20\% have 8k--32k and 5\% have 64k--120k, and outputs have 64--512 tokens. We also use prompt lengths drawn from the Mooncake trace~\cite{qin2024mooncake}, a replay of that trace, and multi-turn sessions over a shared long context (Appendix~\ref{app:setup}). Each rate runs one 120-s arrival window after a 90-s warm-up. Requests of two seeds are pooled, and three on DeepSeek-V4-Flash.

\textbf{SLOs.} The headline SLO is that of \S\ref{sec:bg-slo}, with a TTFT SLO scale of 3 and a P99 TBT of at most 150\,ms. We also score the same runs under three others: the worst-gap variant, the end-to-end deadlines of Niyama and JITServe~\cite{goel2025niyama,zhang2026jitserve}, and TTFT alone. SLO goodput is interpolated between measured rates, and ``at least'' marks a system that stays above 90\% at our highest rate.

\textbf{Baselines.} vLLM~\cite{kwon2023vllm} and Sarathi-Serve (\emph{Sarathi})~\cite{agrawal2024sarathiserve} represent first-come-first-served engines, SGLang's shortest-prefill-first (SPF)~\cite{sglang2026spf} a size-based order, and Medha~\cite{agrawal2024medha} and Niyama~\cite{goel2025niyama} SLO-aware scheduling. Each is configured for our SLO (Appendix~\ref{app:baselines}). The official Sarathi-Serve and Niyama code attain less than our versions (Appendix~\ref{app:official}).

\begin{figure}[t]
\centering
\includegraphics[width=\columnwidth]{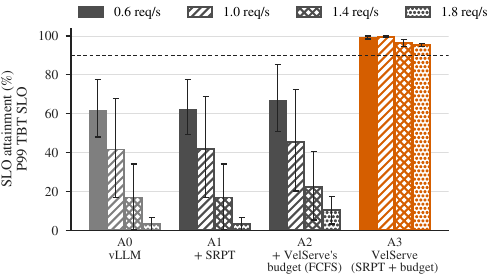}
\caption{Ablation on Qwen3-8B (P99 TBT SLO, two seeds, error bars: seed range). Neither the order (A1) nor the budget (A2) alone exceeds 66\%, and only both together (A3) stay above 90\%.}
\label{fig:ablation}
\end{figure}

\subsection{SLO Goodput under the P99 TBT SLO}
\label{sec:eval-main}

\textbf{\sysname sustains the highest SLO goodput on every model.} On Mixed, the strongest baseline is Medha (Figure~\ref{fig:main}). \sysname sustains 2.4--2.9$\times$ its SLO goodput on Qwen3-8B and DeepSeek-V4-Flash, and at least 4.4$\times$ on DeepSeek-V2-Lite, where it stays above 90\% up to our highest rate. On the replayed Mooncake trace, it sustains 2.7$\times$. The result holds over long runs: in 100 minutes on Qwen3-8B, 95\% of requests met the SLO, and no completed request missed its P99 TBT target.

The baselines fail through the mechanisms of \S\ref{sec:motivation}. FCFS engines and SPF lose first tokens to head-of-line blocking, and Medha and Niyama lose TBT to decode jitters (Table~\ref{tab:overrun}). On DeepSeek-V4-Flash, Medha's iteration target also leaves little room for prefill, so it loses first tokens as well.

\subsection{Robustness to the SLO and Prompt Lengths}
\label{sec:eval-sd}

\textbf{The lead holds across every part of the SLO.} Figure~\ref{fig:sdgap} re-scores the same runs while tightening or relaxing each part of the SLO. \sysname leads under both TBT bounds down to 150\,ms, including the stricter worst-gap bound (Appendix Table~\ref{tab:capacity}). Below that, DeepSeek-V2-Lite reaches a limit set by host stalls of up to about 100\,ms, which no budget can absorb. Looser bounds help the baselines little, because Medha's iteration target limits its prefill and Niyama's longest iterations break even 300\,ms. Configured for 300\,ms (stars), the baselines fill their iterations to the new limit and stay behind. End-to-end deadlines~\cite{zhang2026jitserve,goel2025niyama} forgive pauses for which a stream has banked time, so Niyama's long iterations pass, yet \sysname still leads on all six model--workload pairs (third row, Appendix~\ref{app:e2e}). It also leads on first tokens alone (fourth row), so shortest-first order recovers the prefill time that the margin takes. Only at much tighter TTFT deadlines does Niyama's deadline order serve long prompts sooner (Appendix~\ref{app:details}).

\begin{figure}[t]
\centering
\includegraphics[width=0.82\columnwidth]{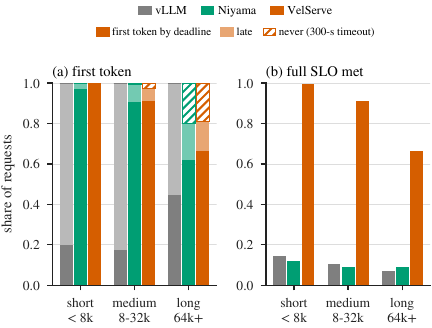}
\caption{Only \sysname meets the full SLO (b) in every class.}
\label{fig:classes}
\end{figure}

\textbf{Prompt lengths change \sysname's SLO goodput, but not its lead.} Mooncake-shaped lengths have far fewer short prompts than Mixed. \sysname still sustains 2.6--4.0$\times$ the strongest baseline, but on DeepSeek-V4-Flash its SLO goodput falls to less than half its value on Mixed. Its losses are first-token losses of the many medium and long prompts, and promoting prompts near their deadline does not recover them (Appendix~\ref{app:order}). With these lengths, the prefill room that the margin leaves limits \sysname more than the order does. Shortest-first order still does not starve long prompts (\S\ref{sec:design-order}). Figure~\ref{fig:classes} breaks a 40-minute run on Qwen3-8B at 1.0\,req/s down by prompt-length class. On first tokens alone, \sysname and Niyama perform similarly in every class (Figure~\ref{fig:classes}a). On the full SLO, only \sysname serves most requests in every class, long prompts included, while the baselines fail most requests in every class (Figure~\ref{fig:classes}b). At this load, no order serves every long prompt: the SLO-aware schedulers each time out on some of the longest (hatched bars), and vLLM avoids timeouts only by making every class wait (Appendix Table~\ref{tab:classes}).

\subsection{Ablation}
\label{sec:eval-ablation}

\textbf{Both remedies are needed.} Figure~\ref{fig:ablation} adds \sysname's order and budget to vLLM, separately and together. Shortest-first order alone (A1) barely helps, because vLLM's large iterations still break the TBT target. \sysname's budget alone (A2) meets the TBT target, but short prompts still wait behind long ones. Neither alone serves two-thirds of requests at any load. Only together (A3) do they keep attainment above 90\%, up to the highest load we ran.

\begin{table}[t]
\caption{SLO attainment (\%) when the margin is measured online, frozen, or moved into Medha (Qwen3-8B, Mixed, P99 TBT SLO, seeds 0--1). Frozen margins keep \sysname's budget rule and order. ``Other workload'': the value measured with Mooncake-shaped lengths.}
\label{tab:margin-main}
\centering\footnotesize
\setlength{\tabcolsep}{3pt}
\begin{tabular}{@{}llrrr@{}}
\toprule
& & \multicolumn{3}{c}{Rate (req/s)}\\
\cmidrule(l){3-5}
Scheduler & Margin & 0.6 & 1.0 & 1.4 \\
\midrule
\sysname & online ($\approx$48\,ms) & 99.3 & 99.6 & 96.4 \\
 & frozen offline (9.4\,ms) & 66.4 & 47.6 & 35.6 \\
 & frozen, other workload (29.3\,ms) & 99.3 & 83.1 & 91.8 \\
\midrule
Medha & none & 100.0 & 57.6 & 25.7 \\
 & 29.3\,ms, other workload & 100.0 & 87.4 & 51.7 \\
 & 48\,ms, \sysname's value & 100.0 & 92.2 & 59.2 \\
\bottomrule
\end{tabular}
\end{table}

\textbf{The margin must be measured on the live workload.} Live calibration avoids the failures of frozen margins, whether they are measured offline or carried over from another workload (Table~\ref{tab:margin-main}). Moved into Medha, the live margin removes Medha's TBT misses but leaves its first-token losses. On DeepSeek-V4-Flash, a hand-tuned fixed margin matches \sysname, whose contribution is to select the margin automatically (Appendix~\ref{app:margin}).

\subsection{Disaggregated Serving}
\label{sec:eval-disagg}

\begin{figure}[t]
\centering
\includegraphics[width=0.88\columnwidth]{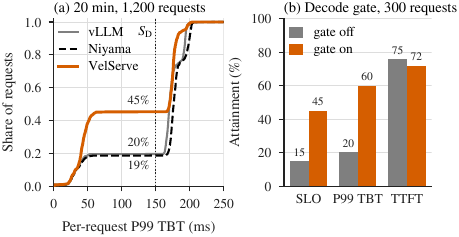}
\caption{(a) The step just above $\SD$ is the join, which \sysname avoids for more requests. (b) The decode gate alone explains this.}
\label{fig:pd}
\end{figure}

\textbf{Setup.} We serve the full DeepSeek-V4-Flash on one node, with four GPUs for prefill and four for decode, each with data-parallel, expert-parallel ranks (Appendix~\ref{app:disagg}). vLLM, Niyama and \sysname share the same prefill configuration and serve the same requests of the Mooncake trace, under the SLO of \S\ref{sec:bg-slo}. Joins keep every system below 90\% attainment at this load, so we compare attainment instead.

\textbf{\sysname meets the SLO for more than twice as many requests} as either baseline (35\%, against 15--16\%). The gain is in TBT: \sysname moves many requests from the join step just above $\SD$ to the fast mode (Figure~\ref{fig:pd}a). Its TTFT attainment stays level with vLLM's and close to Niyama's.

\textbf{The decode gate accounts for the gain.} We toggle only the gate, on the same servers and requests, in a separate window (Figure~\ref{fig:pd}b). With the gate off, \sysname performs like the baselines. With it on, requests gain TBT compliance and none lose it, at a small cost in TTFT. The gate does not shorten joins but makes them rarer, as a P99 target rewards.

\subsection{Scope and Limitations}
\label{sec:eval-limits}
Calibration tracks shifts in the stall distribution with a lag, and under sustained load some of the longest prompts still time out (Appendix~\ref{app:design}). The decode gate helps only for TBT bounds shorter than a join, and we evaluated one P/D allocation on one node and engine (Appendix~\ref{app:disagg}).

\section{Related Work}
\label{sec:related}

\textbf{Iteration-level serving.}
Orca introduced iteration-level scheduling~\cite{yu2022orca}, on which vLLM and SGLang build~\cite{kwon2023vllm,zheng2024sglang}. Sarathi-Serve chunks prefills under a static token budget profiled offline~\cite{agrawal2024sarathiserve}. \sysname keeps this structure but sizes the budget from measured tail latency. Disaggregation~\cite{zhong2024distserve,patel2024splitwise,qin2024mooncake} and multiplexing~\cite{kamath2025podattention,zhu2025nanoflow,gao2026duetserve} separate or overlap the two phases. They move prefill, but a scheduler still decides what each decode step carries and when new requests join. vLLM aligns admission across expert-parallel ranks at a fixed interval~\cite{vllm2026cadence}, which \sysname derives from the SLO.

\textbf{SLO-aware scheduling.}
Medha, Niyama and SLOWeave size each iteration's prefill to a predicted latency or decode deadline~\cite{agrawal2024medha,goel2025niyama,sloweave2026}. Planning to a prediction leaves little room for mispredictions and host stalls, which \sysname absorbs with a measured margin. Li et al.~\cite{li2026uniboost} make the same argument for request order and replace predicted output lengths with running statistics. \sysname pairs its budget with a well-established size-based order~\cite{wu2023fastserve,fu2024ltr,sglang2026spf}. Cascade bounds the TTFT tail with per-request latency budgets~\cite{cascade2026}, and \sysname bounds the TBT tail.

\textbf{Attention architecture and cost.}
GQA, MLA and compressed attention shrink attention cost~\cite{ainslie2023gqa,deepseek2024v2,deepseek2026v4}. Analytical schedulers assume token-linear~\cite{li2025throughput} or KV-load costs~\cite{ao2025fluid}, whereas \sysname tests online which model fits.

\section{Conclusion}
\label{sec:conclusion}

Decode jitters, decode steps that run past the TBT target, arise from how prefill is admitted into decode and delay every stream in the batch. \sysname sizes each iteration against the measured tail of its own latency, recovers the TTFT cost with shortest-first order, and paces admissions under disaggregation by the same principle. It meets strict tail TBT targets for far more requests than prior work.

\bibliographystyle{IEEEtranS}
\bibliography{refs}

@inproceedings{yu2022orca,
  author    = {Yu, Gyeong-In and Jeong, Joo Seong and Kim, Geon-Woo and Kim, Soojeong and Chun, Byung-Gon},
  title     = {Orca: A Distributed Serving System for Transformer-Based Generative Models},
  booktitle = {16th USENIX Symposium on Operating Systems Design and Implementation (OSDI)},
  pages     = {521--538},
  year      = {2022}
}

@inproceedings{kwon2023vllm,
  author    = {Kwon, Woosuk and Li, Zhuohan and Zhuang, Siyuan and Sheng, Ying and Zheng, Lianmin and Yu, Cody Hao and Gonzalez, Joseph E. and Zhang, Hao and Stoica, Ion},
  title     = {Efficient Memory Management for Large Language Model Serving with {PagedAttention}},
  booktitle = {Proceedings of the 29th Symposium on Operating Systems Principles (SOSP)},
  pages     = {611--626},
  year      = {2023}
}

@inproceedings{zheng2024sglang,
  author    = {Zheng, Lianmin and Yin, Liangsheng and Xie, Zhiqiang and Sun, Chuyue and Huang, Jeff and Yu, Cody Hao and Cao, Shiyi and Kozyrakis, Christos and Stoica, Ion and Gonzalez, Joseph E. and Barrett, Clark and Sheng, Ying},
  title     = {{SGLang}: Efficient Execution of Structured Language Model Programs},
  booktitle = {Advances in Neural Information Processing Systems (NeurIPS)},
  year      = {2024}
}

@article{agrawal2023sarathi,
  author  = {Agrawal, Amey and Panwar, Ashish and Mohan, Jayashree and Kwatra, Nipun and Gulavani, Bhargav S. and Ramjee, Ramachandran},
  title   = {{SARATHI}: Efficient {LLM} Inference by Piggybacking Decodes with Chunked Prefills},
  journal = {arXiv preprint arXiv:2308.16369},
  year    = {2023}
}

@inproceedings{agrawal2024sarathiserve,
  author    = {Agrawal, Amey and Kedia, Nitin and Panwar, Ashish and Mohan, Jayashree and Kwatra, Nipun and Gulavani, Bhargav S. and Tumanov, Alexey and Ramjee, Ramachandran},
  title     = {Taming Throughput-Latency Tradeoff in {LLM} Inference with {Sarathi-Serve}},
  booktitle = {18th USENIX Symposium on Operating Systems Design and Implementation (OSDI)},
  pages     = {117--134},
  year      = {2024}
}

@inproceedings{zhong2024distserve,
  author    = {Zhong, Yinmin and Liu, Shengyu and Chen, Junda and Hu, Jianbo and Zhu, Yibo and Liu, Xuanzhe and Jin, Xin and Zhang, Hao},
  title     = {{DistServe}: Disaggregating Prefill and Decoding for Goodput-optimized Large Language Model Serving},
  booktitle = {18th USENIX Symposium on Operating Systems Design and Implementation (OSDI)},
  pages     = {193--210},
  year      = {2024}
}

@inproceedings{patel2024splitwise,
  author    = {Patel, Pratyush and Choukse, Esha and Zhang, Chaojie and Shah, Aashaka and Goiri, {\'I}{\~n}igo and Maleki, Saeed and Bianchini, Ricardo},
  title     = {Splitwise: Efficient Generative {LLM} Inference Using Phase Splitting},
  booktitle = {51st ACM/IEEE International Symposium on Computer Architecture (ISCA)},
  pages     = {118--132},
  doi       = {10.1109/ISCA59077.2024.00019},
  year      = {2024}
}

@inproceedings{qin2024mooncake,
  author    = {Qin, Ruoyu and Li, Zheming and He, Weiran and Cui, Jialei and Ren, Feng and Zhang, Mingxing and Wu, Yongwei and Zheng, Weimin and Xu, Xinran},
  title     = {Mooncake: Trading More Storage for Less Computation --- A {KVCache}-centric Architecture for Serving {LLM} Chatbot},
  booktitle = {23rd USENIX Conference on File and Storage Technologies (FAST)},
  pages     = {155--170},
  url       = {https://www.usenix.org/conference/fast25/presentation/qin},
  note      = {arXiv:2407.00079},
  year      = {2025}
}

@inproceedings{kamath2025podattention,
  author    = {Kamath, Aditya K. and Prabhu, Ramya and Mohan, Jayashree and Peter, Simon and Ramjee, Ramachandran and Panwar, Ashish},
  title     = {{POD-Attention}: Unlocking Full Prefill-Decode Overlap for Faster {LLM} Inference},
  booktitle = {Proceedings of the 30th ACM International Conference on Architectural Support for Programming Languages and Operating Systems (ASPLOS)},
  year      = {2025}
}

@inproceedings{zhu2025nanoflow,
  author    = {Zhu, Kan and Gao, Yufei and Zhao, Yilong and Zhao, Liangyu
          and Zuo, Gefei and Gu, Yile and Xie, Dedong and Tang, Tian
          and Xu, Qinyu and Ye, Zihao and Kamahori, Keisuke
          and Lin, Chien-Yu and Wang, Ziren and Wang, Stephanie
          and Krishnamurthy, Arvind and Kasikci, Baris},
  title     = {{NanoFlow}: Towards Optimal Large Language Model Serving Throughput},
  booktitle = {19th USENIX Symposium on Operating Systems Design and Implementation (OSDI)},
  year      = {2025}
}

@article{agrawal2024medha,
  author  = {Agrawal, Amey and Qiu, Haoran and Chen, Junda and Goiri, {\'I}{\~n}igo and Zhang, Chaojie and Shahid, Rayyan and Ramjee, Ramachandran and Tumanov, Alexey and Choukse, Esha},
  title   = {No Request Left Behind: Tackling Heterogeneity in Long-Context {LLM} Inference with {Medha}},
  journal = {arXiv preprint arXiv:2409.17264},
  year    = {2024}
}

@inproceedings{goel2025niyama,
  author    = {Goel, Kanishk and Mohan, Jayashree and Kwatra, Nipun and Anupindi, Ravi Shreyas and Ramjee, Ramachandran},
  title     = {{QoServe}: Breaking the Silos of {LLM} Inference Serving},
  booktitle = {Proceedings of the 31st ACM International Conference on Architectural Support for Programming Languages and Operating Systems (ASPLOS)},
  doi       = {10.1145/3779212.3790206},
  note      = {Preprint as \emph{Niyama: Breaking the Silos of LLM Inference Serving}, arXiv:2503.22562},
  year      = {2026}
}

@article{chen2025slosserve,
  author  = {Chen, Siyuan and Jia, Zhipeng and Khan, Samira and Krishnamurthy, Arvind and Gibbons, Phillip B.},
  title   = {{SLOs-Serve}: Optimized Serving of Multi-{SLO} {LLMs}},
  journal = {arXiv preprint arXiv:2504.08784},
  year    = {2025}
}

@article{sloweave2026,
  author  = {Song, Siyu and Bai, Qi and Hao, Jinbo and Li, Kai and Wang, Chenchen and Sun, Jiayu},
  title   = {Deadline-Aware Adaptive Prefill Chunking for Efficient Large Language Model Serving},
  journal = {arXiv preprint arXiv:2609.07883},
  year    = {2026}
}

@inproceedings{wu2023fastserve,
  author    = {Wu, Bingyang and Zhong, Yinmin and Zhang, Zili and Liu, Shengyu and Liu, Fangyue and Sun, Yuanhang and Huang, Gang and Liu, Xuanzhe and Jin, Xin},
  title     = {{FastServe}: Iteration-Level Preemptive Scheduling for Large Language Model Inference},
  booktitle = {23rd USENIX Symposium on Networked Systems Design and Implementation (NSDI)},
  pages     = {57--74},
  note      = {Preprint: arXiv:2305.05920},
  year      = {2026}
}

@inproceedings{fu2024ltr,
  author    = {Fu, Yichao and Zhu, Siqi and Su, Runlong and Qiao, Aurick and Stoica, Ion and Zhang, Hao},
  title     = {Efficient {LLM} Scheduling by Learning to Rank},
  booktitle = {Advances in Neural Information Processing Systems (NeurIPS)},
  year      = {2024}
}

@article{cascade2026,
  author  = {Adnan, Muhammad and Mahapatra, Rohan and Nair, Prashant J. and Berger, Daniel and Zardoshti, Pantea and Fonseca, Rodrigo and Choukse, Esha},
  title   = {Cascade: Exploiting {SLO}-Aware Latency Budget for Fair and High Goodput {LLM} Inference Serving},
  journal = {arXiv preprint arXiv:2608.06557},
  year    = {2026}
}

@misc{sglang2026spf,
  author       = {{SGLang Contributors}},
  title        = {[Scheduler] Add shortest-prefill-first scheduling (pull request \#40024, merged 2026-09-18)},
  howpublished = {\url{https://github.com/sgl-project/sglang/pull/40024}},
  year         = {2026}
}

@inproceedings{ainslie2023gqa,
  author    = {Ainslie, Joshua and Lee-Thorp, James and de Jong, Michiel and Zemlyanskiy, Yury and Lebr{\'o}n, Federico and Sanghai, Sumit},
  title     = {{GQA}: Training Generalized Multi-Query Transformer Models from Multi-Head Checkpoints},
  booktitle = {Proceedings of the 2023 Conference on Empirical Methods in Natural Language Processing (EMNLP)},
  year      = {2023}
}

@article{deepseek2024v2,
  author  = {{DeepSeek-AI}},
  title   = {{DeepSeek-V2}: A Strong, Economical, and Efficient Mixture-of-Experts Language Model},
  journal = {arXiv preprint arXiv:2405.04434},
  year    = {2024}
}

@article{deepseek2026v4,
  author  = {{DeepSeek-AI}},
  title   = {{DeepSeek-V4}: Towards Highly Efficient Million-Token Context Intelligence},
  journal = {arXiv preprint arXiv:2606.19348},
  year    = {2026}
}

@article{li2025throughput,
  author  = {Dai, J. G. and Deng, Tianze and Li, Yueying and Peng, Tianyi},
  title   = {Throughput-Optimal Scheduling Algorithms for {LLM} Inference and {AI} Agents},
  journal = {arXiv preprint arXiv:2504.07347},
  year    = {2025}
}

@article{ao2025fluid,
  author  = {Ao, Ruicheng and Luo, Gan and Simchi-Levi, David and Wang, Xinshang},
  title   = {Optimizing {LLM} Inference: Fluid-Guided Online Scheduling with Memory Constraints},
  journal = {arXiv preprint arXiv:2504.11320},
  year    = {2025}
}

@article{schrage1968,
  author  = {Schrage, Linus},
  title   = {A Proof of the Optimality of the Shortest Remaining Processing Time Discipline},
  journal = {Operations Research},
  volume  = {16},
  number  = {3},
  pages   = {687--690},
  year    = {1968}
}

@article{smith1978,
  author  = {Smith, Donald R.},
  title   = {A New Proof of the Optimality of the Shortest Remaining Processing Time Discipline},
  journal = {Operations Research},
  volume  = {26},
  number  = {1},
  pages   = {197--199},
  year    = {1978}
}

@article{harcholbalter2002,
  author  = {Harchol-Balter, Mor and Sigman, Karl and Wierman, Adam},
  title   = {Asymptotic Convergence of Scheduling Policies with Respect to Slowdown},
  journal = {Performance Evaluation},
  volume  = {49},
  number  = {1--4},
  pages   = {241--256},
  year    = {2002}
}

@article{moore1968,
  author  = {Moore, J. Michael},
  title   = {An n Job, One Machine Sequencing Algorithm for Minimizing the Number of Tardy Jobs},
  journal = {Management Science},
  volume  = {15},
  number  = {1},
  pages   = {102--109},
  year    = {1968}
}

@article{dean2013tail,
  author  = {Dean, Jeffrey and Barroso, Luiz Andr{\'e}},
  title   = {The Tail at Scale},
  journal = {Communications of the ACM},
  volume  = {56},
  number  = {2},
  pages   = {74--80},
  year    = {2013}
}

@article{brysbaert2019reading,
  author  = {Brysbaert, Marc},
  title   = {How Many Words Do We Read per Minute? {A} Review and Meta-Analysis of Reading Rate},
  journal = {Journal of Memory and Language},
  volume  = {109},
  pages   = {104047},
  year    = {2019}
}

@inproceedings{bender1998stretch,
  author    = {Bender, Michael A. and Chakrabarti, Soumen and Muthukrishnan, S.},
  title     = {Flow and Stretch Metrics for Scheduling Continuous Job Streams},
  booktitle = {Proceedings of the ACM-SIAM Symposium on Discrete Algorithms (SODA)},
  pages     = {270--279},
  year      = {1998}
}

@inproceedings{zhang2026jitserve,
  author    = {Zhang, Wei and Wu, Zhiyu and Mu, Yi and Ning, Rui and Liu, Banruo and Sarda, Nikhil and Lee, Myungjin and Lai, Fan},
  title     = {{JITServe}: {SLO}-aware {LLM} Serving with Imprecise Request Information},
  booktitle = {23rd USENIX Symposium on Networked Systems Design and Implementation (NSDI)},
  pages     = {825--848},
  year      = {2026}
}

@misc{vllm2026dp,
  author       = {{vLLM Team}},
  title        = {Data Parallel Deployment},
  year         = {2026},
  howpublished = {\url{https://docs.vllm.ai/en/latest/serving/data_parallel_deployment/}},
  note         = {Accessed October 2026}
}

@misc{vllm2026cadence,
  author       = {{vLLM Team}},
  title        = {{vLLM} x {AgentX}: Optimizing for Real-World Agentic Serving},
  year         = {2026},
  howpublished = {\url{https://github.com/vllm-project/vllm-project.github.io/blob/main/_posts/2026-09-08-vllm-agentx.md}},
  note         = {Blog post, September 8, 2026; section ``Align DEP prefill schedule cadence''}
}

@inproceedings{dewitt1984,
  author    = {DeWitt, David J. and Katz, Randy H. and Olken, Frank and Shapiro, Leonard D. and Stonebraker, Michael R. and Wood, David A.},
  title     = {Implementation Techniques for Main Memory Database Systems},
  booktitle = {Proceedings of the 1984 ACM SIGMOD International Conference on Management of Data},
  pages     = {1--8},
  year      = {1984}
}

@inproceedings{maas2015trashday,
  author    = {Maas, Martin and Harris, Tim and Asanovi{\'c}, Krste and Kubiatowicz, John},
  title     = {Trash Day: Coordinating Garbage Collection in Distributed Systems},
  booktitle = {15th Workshop on Hot Topics in Operating Systems (HotOS XV)},
  year      = {2015}
}

@article{kao1997deadline,
  author  = {Kao, Ben and Garcia-Molina, Hector},
  title   = {Deadline Assignment in a Distributed Soft Real-Time System},
  journal = {IEEE Transactions on Parallel and Distributed Systems},
  volume  = {8},
  number  = {12},
  pages   = {1268--1274},
  year    = {1997}
}

@inproceedings{gao2026duetserve,
  author    = {Gao, Lei and Jiang, Chaoyi and Entezari Zarch, Hossein and Wong, Daniel and Hill, Mark D. and Annavaram, Murali},
  title     = {{DuetServe}: Harmonizing Prefill and Decode for {LLM} Serving via Adaptive {GPU} Multiplexing},
  booktitle = {Proceedings of the 43rd International Conference on Machine Learning (ICML)},
  series    = {Proceedings of Machine Learning Research},
  volume    = {306},
  pages     = {33270--33284},
  url       = {https://proceedings.mlr.press/v306/gao26m.html},
  note      = {arXiv:2511.04791},
  year      = {2026}
}

@inproceedings{li2026uniboost,
  author    = {Li, Yueying and Chen, Yuanfan and Chen, Jiayang and Choukse, Esha and Qiu, Haoran and Suh, G. Edward and Fonseca, Rodrigo and Scully, Ziv and Gupta, Udit},
  title     = {Beyond Prediction: Tail-Aware Scheduling for {LLM} Inference},
  booktitle = {Proceedings of the 43rd International Conference on Machine Learning (ICML)},
  series    = {Proceedings of Machine Learning Research},
  volume    = {306},
  note      = {arXiv:2606.18431},
  year      = {2026}
}

@article{liu2024andes,
  author  = {Liu, Jiachen and Chung, Jae-Won and Wu, Zhiyu and Lai, Fan and Lee, Myungjin and Chowdhury, Mosharaf},
  title   = {Andes: Defining and Enhancing Quality-of-Experience in {LLM}-Based Text Streaming Services},
  journal = {arXiv preprint arXiv:2404.16283},
  year    = {2024}
}

@article{wang2024revisiting,
  author  = {Wang, Zhibin and Li, Shipeng and Zhou, Yuhang and Li, Xue and Zhang, Zhonghui and Cam-Tu, Nguyen and Gu, Rong and Tian, Chen and Chen, Guihai and Zhong, Sheng},
  title   = {Revisiting Service Level Objectives and System Level Metrics in Large Language Model Serving},
  journal = {arXiv preprint arXiv:2410.14257},
  year    = {2024}
}

@article{kamahori2026voxserve,
  author  = {Kamahori, Keisuke and Lee, Wei-Tzu and Jha, Atindra and Kadekodi, Rohan and Wang, Stephanie and Krishnamurthy, Arvind and Kasikci, Baris},
  title   = {{VoxServe}: Streaming-Centric Serving System for Speech Language Models},
  journal = {arXiv preprint arXiv:2602.00269},
  year    = {2026}
}

\clearpage
\appendices
\renewcommand{\theHsection}{appendix.\Alph{section}}
\makeatletter
\@addtoreset{figure}{section}\@addtoreset{table}{section}\@addtoreset{algorithm}{section}\@addtoreset{equation}{section}
\makeatother
\renewcommand{\thefigure}{\thesection.\arabic{figure}}
\renewcommand{\thetable}{\thesection.\arabic{table}}
\renewcommand{\thealgorithm}{\thesection.\arabic{algorithm}}
\renewcommand{\theequation}{\thesection.\arabic{equation}}
\renewcommand{\theHfigure}{\theHsection.\arabic{figure}}
\renewcommand{\theHtable}{\theHsection.\arabic{table}}
\renewcommand{\theHalgorithm}{\theHsection.\arabic{algorithm}}
\renewcommand{\theHequation}{\theHsection.\arabic{equation}}
\makeatletter
\setlength{\@fptop}{0pt}\setlength{\@fpsep}{14pt}\setlength{\@fpbot}{0pt plus 1fil}
\setlength{\@dblfptop}{0pt}\setlength{\@dblfpsep}{14pt}\setlength{\@dblfpbot}{0pt plus 1fil}
\makeatother
\renewcommand{\topfraction}{0.9}\renewcommand{\dbltopfraction}{0.9}\renewcommand{\textfraction}{0.07}
\renewcommand{\floatpagefraction}{0.85}\renewcommand{\dblfloatpagefraction}{0.85}
\setcounter{topnumber}{3}\setcounter{dbltopnumber}{2}\setcounter{totalnumber}{4}

\noindent Section, figure, table and equation numbers without a letter refer to the main paper. Unless stated otherwise, results use the headline SLO: the TTFT deadline and a per-request P99 TBT of at most 150\,ms.

\section{Supporting Analysis}
\label{app:analysis}

This section proves the claims of \S\ref{sec:design} under a simple model of an iteration. The scheduling arguments are classical. What matters is the condition on iteration cost under which each one holds.

\begin{proposition}[Tail-sized budget]\label{prop:tail}
Suppose a request meets its TBT target if at most $k$ of its gaps exceed $\SD-\delta$. Suppose an iteration at budget $\Nstar$, with decode count and contexts within the reference load, lasts at most $\hat t(\Nstar)+\epsilon$, and $\epsilon$ has a stationary marginal distribution. With $\eta=\beta(1-\alpha)(k+1)/\bar o$, the expected fraction of requests that violate the TBT target is at most $\beta(1-\alpha)$.
\end{proposition}

\begin{proposition}[Shortest remaining first]\label{prop:srpt}
Under a linear, context-free iteration cost and a given decode trajectory $(b_s)$, prefill is a single preemptive server that offers $\Nstar-w\,b_s$ tokens in iteration $s$. Filling each iteration in order of fewest remaining prompt tokens maximizes the number of completed prefills at every iteration boundary and minimizes the sum of TTFTs among policies that respect the budget, for any constant decode weight $w$.
\end{proposition}

\begin{proposition}[Weighted token budget]\label{prop:budget}
If iteration time is linear and context-free ($g=0$, constant $\cp,\cd$), the largest prefill an iteration with $b$ decodes can carry within $\SD-\delta$ is $\Nstar-w\,b$, with $w=\cd/\cp$ and $\Nstar=(\SD-\delta-a)/\cp$, independent of load, request mix and context. A plain token budget, which counts a decode as one token, matches it for every $b$ iff $w=1$. If $g>0$, no context-blind budget does.
\end{proposition}

\begin{proposition}[Context-dependent cost]\label{prop:nu}
If a prefill token at offset $k$ costs $\cp(1+\gamma k)$, ordering by remaining tokens has total completion time at most $\nu=1+\gamma k_{\max}$ times the optimum (common release).
\end{proposition}

\textbf{Model.} Iteration $s$ carries $b_s$ decodes and $p_s$ prefill tokens, of which $x_{r,s}$ belong to request $r$. It lasts $t_s=a+\cd b_s+\cp p_s+g\sum_i C_ik_i+\epsilon_s$ (Eq.~\ref{eq:cost}), where $\epsilon_s$ is the part of its time that the model does not explain. Every running decode receives one token per iteration, and the TBT SLO asks for $t_s\le\SD-\delta$.

\textbf{Proof of Proposition~\ref{prop:tail}.} An iteration runs past $\SD-\delta$ only if $\epsilon_s$ exceeds the margin $Q$, which happens with probability $\eta$. Let $N$ count the late iterations that a request waits on. A request with $o$ output tokens waits on $o$ iterations, so $\mathbb{E}[N]\le o\eta$ by linearity of expectation, with no independence assumption. Markov's inequality gives $\Pr[N>k]\le\mathbb{E}[N]/(k+1)$, and averaging over requests gives Eq.~(\ref{eq:tail}).

One caveat applies. Here $\eta$ is the rate that requests see, so an iteration with $b$ decodes counts $b$ times. The calibrator measures the rate over iterations, and requests see a higher rate when overruns fall on large batches (Figure~\ref{fig:tail}a). The bound therefore holds only if the calibrated $Q$ achieves $\eta$ in this per-token sense. Our runs are consistent with it.

\textbf{A P99 target tolerates any mix of output lengths.} With $k=0$, the bound is the union bound, which gives the worst-gap rule. A per-request P99 target over $n$ gaps allows $k=n-1-\lfloor0.99(n-1)\rfloor$ late gaps, so its allowance grows with the response. Since $n/(k+1)<100$ for every $n$, a rate $\eta\le\beta(1-\alpha)/100$ meets the P99 target whatever the output lengths. A bound on every gap has no such property, because its tolerable rate falls as $1/n$.

\textbf{Rare jitters reach most long answers.} If each step were late independently with probability $q$, a request with $o$ tokens would avoid every late step with probability $(1-q)^o$. At $q=1\%$, a 300-token answer meets at least one late step with probability $1-0.99^{300}\approx0.95$. A margin at the 99th percentile of iteration time is therefore far too loose for long answers. For $\alpha=0.99$, $\beta=0.5$, $\bar o=300$ and $k=0$, as in all our runs, Eq.~(\ref{eq:tail}) asks for $\eta\approx1.7\times10^{-5}$.

\textbf{Proof of Proposition~\ref{prop:srpt}.} The budget limits how many prefill tokens an iteration carries, not whose they are. Every policy that fills the budget while prompts wait therefore has the same remaining work, and the same iteration lengths, at every boundary. The classical exchange argument for shortest-remaining-processing-time (SRPT) scheduling~\cite{schrage1968,smith1978} then applies. Serving a token of the request with less work left never delays the earlier of two completions, so shortest-first completes the most prefills by every boundary. The sum of TTFTs adds up the unfinished requests over iterations, so it is minimal too. The decode weight $w$ changes how much prefill fits, not how two prefills compare. The result is conditional on the decode trajectory, which a different order changes, so we also evaluate the order by experiment.

\textbf{Proof of Proposition~\ref{prop:budget}.} The TBT limit $a+\cd b+\cp p\le\SD-\delta$ is the same as $p+wb\le\Nstar$, and carrying less prefill delays first tokens without shortening any TBT. A plain budget $N$ allows $N-b$ prefill tokens, which differs from the safe amount by $(\Nstar-N)+(1-w)b$. This is zero for every $b$ only if $w=1$ and $N=\Nstar$. Otherwise a plain budget must be sized for the worst batch and leaves prefill room unused in all others. If cost depends on context ($g>0$), so does the safe amount, and no context-blind budget is exact.

\textbf{Proof of Proposition~\ref{prop:nu}.} With context-dependent cost, a request with $R$ tokens left needs between $\cp R$ and $\nu\cp R$ of time. Token order is optimal when work is counted in tokens. Since each request's time is within a factor $\nu$ of its token count, token order is at most $\nu$ times worse than the best order counted in time.

\textbf{How large is $\nu$?} On compressed attention, a token's cost barely grows with its offset, and $\nu$ stays near 1 up to 128k-token contexts. Under MLA and GQA, it reaches about 6 at 128k. \sysname checks two conditions online (\S\ref{sec:design-cost}). The first is $\nu\le1.25$ at the 95th-percentile context. The second is $|\cd-\cp|\,b_{90}\le0.05\,\SD$, where $b_{90}$ is the 90th-percentile decode count, which keeps a plain token budget accurate. On DeepSeek-V4-Flash, the first always holds, but the second fails in about half of the calibrations, so \sysname uses the weighted budget. On MLA and GQA, the first always fails, so \sysname sizes by time.

\textbf{How long do long prompts wait?} Let $\mu$ be the prefill rate when the budget is full, and let requests shorter than $x$ bring at most $\sigma+\rho_{<x}\mu\tau$ tokens of work in any window of length $\tau$. Under SRPT, a request of size $x$ is served whenever no shorter work waits. If $V$ tokens of shorter work are queued at its arrival, it finishes within $(V+x+\sigma)/(\mu(1-\rho_{<x}))$. Its deadline is about $\kappa x/\mu_{\mathrm{solo}}$, so for large $x$ it is met only if shorter requests use at most about $1-1/\kappa$ of the prefill capacity, two thirds at $\kappa=3$. Beyond that load, no order helps much. For the largest requests, every policy that never idles has a slowdown near $1/(1-\rho)$~\cite{harcholbalter2002}. This is why medium and long prompts wait long with Mooncake-shaped lengths (\S\ref{sec:eval-sd}).

\textbf{Why late requests move to the back.} If all requests arrive together, their deadlines grow with prompt length, so earliest-deadline order is shortest-first order. The Moore--Hodgson algorithm~\cite{moore1968} then maximizes the number of on-time requests, and in this order it reduces to one rule: serve the shortest first, and skip any request that would be late. When requests arrive over time and can be paused, no algorithm is guaranteed to stay within a constant factor of the best, so we evaluate this rule by experiment.

\section{Design Details}
\label{app:design}

\begin{algorithm}[t]
\caption{Budget calibration (every 500 iterations)}
\label{alg:calib}
\small\raggedright
\begin{algorithmic}[1]
\REQUIRE iteration log; SLO $\SD$; target $\alpha$; floor $Q_0$
\STATE fit $\theta=(a,\cp,\cd,g)$ of Eq.~(\ref{eq:cost}) on the last 8k iterations (Huber loss, offline fit as prior)
\STATE $\epsilon_s\gets t_s-\hat t_\theta(s)$ for the last 200k iterations with decodes
\STATE $\eta\gets\beta(1-\alpha)(k+1)/\bar o$ \COMMENT{$\bar o$: mean output length; $k=0$ in all runs}
\STATE $x_{\min}\gets\SD-\delta-\hat t_\theta(N_{\min};\text{decaying reference})$; $f_{\min}\gets$ share of $\epsilon>x_{\min}$ \COMMENT{stalls no budget absorbs}
\STATE $Q\gets$ rank statistic: $\epsilon$ exceeded by a share $\eta+f_{\min}$ of the window
\IF{window $<3/\eta$ samples} \STATE $Q\gets\max(Q,Q_0)$ \ENDIF
\STATE reference load $(b_{\mathrm{ref}},k_{\mathrm{ref}})\gets$ max of its old value and the 90th-pct.\ decode count, 95th-pct.\ chunk context
\STATE decaying reference $\gets$ max of the same quantiles and its old value $\times\,2^{-500/32\mathrm{k}}$
\STATE mode $\gets$ \textsc{token} if $\nu(k_{\mathrm{ref}})\le1.25$ and $|\cd-\cp|\,b_{\mathrm{ref}}\le0.05\,\SD$, else \textsc{time}
\STATE $N\gets\max\{N:\hat t_\theta(N;b_{\mathrm{ref}},k_{\mathrm{ref}})+Q\le\SD-\delta\}$
\IF{$N<\Nstar$ \OR $N>1.05\,\Nstar$ in two consecutive calls} \STATE $\Nstar\gets N$ \ENDIF
\end{algorithmic}
\end{algorithm}

\begin{algorithm}[t]
\caption{Planning one iteration}
\label{alg:plan}
\small\raggedright
\begin{algorithmic}[1]
\REQUIRE running decodes $B$ ($b=|B|$); prompts $W$ with deadlines $D_r$; $\theta$, $Q$, $\Nstar$, mode
\FORALL{$r\in W$}
\STATE $R_r\gets$ remaining uncached prompt tokens (\textsc{token}) or their predicted time $\hat t_\theta$ (\textsc{time})
\STATE mark $r$ late if, served alone from now, it would finish after $D_r$
\ENDFOR
\STATE order $\gets$ feasible prompts by $R_r$ ascending, then late prompts by arrival
\IF{$b=0$} \STATE $U\gets$ vLLM's maximum \COMMENT{no TBT to protect}
\ELSIF{mode $=$ \textsc{token}} \STATE $U\gets\Nstar-b$
\ELSE \STATE $U\gets\max\{P:\hat t_\theta(B,\,\mathrm{order}_{\le P})+Q\le\SD-\delta\}$ \COMMENT{bisection; $\mathrm{order}_{\le P}$: first $P$ tokens}
\ENDIF
\STATE walk the order, giving each prompt $\min(\text{remaining},\text{left of }U)$ tokens (at most 6 chunks)
\STATE \textbf{return} queue order and per-request chunk caps to vLLM
\end{algorithmic}
\end{algorithm}

Algorithm~\ref{alg:calib} gives the calibration that runs every 500 iterations, and Algorithm~\ref{alg:plan} plans one iteration.

\textbf{The margin comes from \sysname's own iterations.} The prediction error depends on what each iteration contains, and the budget decides that. A margin measured under another policy therefore describes iterations that the scheduler no longer runs (Appendix~\ref{app:overrun}). \sysname refits its cost model on its recent iterations and takes $Q$ from their residuals, including those of iterations near the budget, where the bound binds.

\textbf{The margin is a rank statistic.} The residuals form a light bulk plus rare clusters of host stalls, which a fitted distribution would miss. $Q$ is therefore a high-ranked residual over the last 200k iterations. Some stalls would break $\SD$ even in the smallest iteration, so no budget can absorb them. \sysname sets these aside and counts their share in the rank, so $Q$ sits just below them. Until the window holds enough residuals, a starting margin $Q_0$, measured offline, serves as a floor. It rarely binds in our runs.

\textbf{Calibration adapts with a lag.} The budget drops at once but rises only after two consecutive proposals. Its reference load keeps the heaviest load seen, so a quiet period never inflates it. The cut-off for irreducible stalls uses a separate reference that decays over time, so one burst of heavy load cannot hold it down for the rest of a run. A sudden burst of deeper stalls can still cost some attainment before $Q$ catches up. If the budget reaches its minimum, \sysname flags the TBT SLO as infeasible instead of silently missing it.

\textbf{Cache-aware sizing.} With prefix caching, vLLM applies a waiting request's cache hit only when it admits the request. A follow-up turn with a long cached prefix would therefore be sized as a full prompt. \sysname looks up each waiting request's cached prefix with the engine's own block-hash lookup and sizes the request by its uncached work. The TTFT deadline stays defined on the full prompt.

\textbf{Implementation.} \sysname is a plugin for vLLM's V1 scheduler, with about 1k lines of Python. It reorders the queues by its plan and caps each request's chunk through an existing vLLM hook. Batching, KV allocation, CUDA graphs and asynchronous scheduling are unchanged, and a planning error falls back to the stock scheduler. Planning takes under a millisecond per iteration even at the 99th percentile, and asynchronous scheduling overlaps it with the GPU. Refits run in a background thread.

\section{Setup, Metrics and Baselines}
\label{app:setup}

\begin{table}[t]
\caption{Setup of the colocated experiments. Appendix~\ref{app:disagg} describes the disaggregated deployment.}
\label{tab:setup-full}
\centering
\footnotesize
\setlength{\tabcolsep}{3pt}
\begin{tabular}{@{}p{1.45cm}p{6.3cm}@{}}
\toprule
GPU & 1$\times$ NVIDIA H100 NVL (94\,GB HBM3, 132 SMs, 400\,W cap) \\
Software & vLLM 0.30.0 (V1), CUDA 13.0, PyTorch 2.13; CUDA graphs up to 2304-token iterations, asynchronous scheduling, $\le$256 running requests, 90\% of memory for weights and KV \\
\midrule
\multicolumn{2}{@{}l}{\emph{Models} (attention; layers; weights; KV per token)} \\
GQA & Qwen3-8B; 36; real weights, YaRN-extended to 128k; $\approx$144\,KB \\
MLA & DeepSeek-V2-Lite; 27; real weights; $\approx$30\,KB \\
CSA/HCA & DeepSeek-V4-Flash; 43; real weights of routed experts 0--63 of 256 per layer, router sliced to them (the per-GPU expert footprint of a 4-way expert-parallel shard); $\approx$10\,KB \\
\midrule
Workloads & \emph{Mixed}: Poisson arrivals; prompts 128--2k tokens (75\%), 8k--32k (20\%) and 64k--120k (5\%); outputs 64--512 tokens. \emph{Mooncake-shaped}: Poisson arrivals; prompt and output lengths drawn from the Mooncake conversation trace~\cite{qin2024mooncake}. \emph{Mooncake trace} (CSA only): 150 consecutive requests of the same trace, replayed and time-scaled. \emph{Multi-turn}: sessions of 3--6 turns over a 16k--96k-token document, plus short background requests. \emph{Agent trace} (CSA): Mooncake tool-agent requests with the prefixes they share. Prefix caching is on for the last two only \\
Protocol & One server per system and model; 90-s warm-up at the highest rate, then one 120-s arrival window per rate (trace: 150\,s), in ascending order. Two seeds by default, three for the main DeepSeek-V4-Flash comparisons; four slices of the trace \\
\bottomrule
\end{tabular}
\end{table}

Table~\ref{tab:setup-full} gives the setup of the colocated experiments. The full DeepSeek-V4-Flash checkpoint does not fit on one GPU. We therefore keep the experts that one GPU holds under 4-way expert parallelism and slice the router to them. Per-token compute, weight traffic, power draw and routing skew then match the real model. The generated text does not, so output lengths are fixed per request.

\textbf{Metrics.} An open-loop client records each request's arrival and the time of every output token. A request meets the SLO if its first token arrives within $\max(1\,\mathrm{s},\,3\,\tsolo(P))$ and the 99th percentile of its own gaps is at most $\SD$. Tokens delivered together count as one gap followed by zero gaps. The \emph{worst-gap} variant requires every gap to be at most $\SD$, and Appendix~\ref{app:e2e} defines the end-to-end variant.

\textbf{Scoring.} Every request that arrives in the window counts, and a request that fails or times out after 300\,s is a miss. Attainment pools the requests of all seeds. SLO goodput is the rate at which attainment first falls below 90\%, interpolated between rates. If a system stays above 90\% at our highest rate, its SLO goodput is a lower bound. A system that falls below 50\% at two consecutive rates is not run at higher rates, which count as zero. Every alternative SLO re-scores the same stored per-token logs. The MLA and GQA grids first ran with a low-load warm-up. We re-ran the 13 cells whose misses fell into at most two shared stalls, a rule fixed in advance, and 3 of them changed.

\textbf{Baselines.} All systems run on the same vLLM build and flags, and each is configured for the SLO it is scored against (Appendix~\ref{app:baselines}).
(1)~\emph{vLLM}: first-come-first-served chunked prefill with its default 8192-token budget.
(2)~\emph{Sarathi}: vLLM with a static budget sized for the SLO, as in Sarathi-Serve~\cite{agrawal2024sarathiserve}.
(3)~\emph{SPF}: SGLang's shortest-prefill-first~\cite{sglang2026spf} at the static budget.
(4)~\emph{Niyama}: our reimplementation of Niyama~\cite{goel2025niyama}, with an automatically measured margin.
(5)~\emph{Medha}: our implementation of Medha's single-replica scheduler~\cite{agrawal2024medha}, with no margin.
(6)~\emph{\sysname}, configured by the SLO, $\alpha=0.99$ and a starting margin $Q_0$. The DeepSeek-V4-Flash runs predate the decaying cut-off reference of Algorithm~\ref{alg:calib}. A check run with it settled at the same margin.

\textbf{Measurement notes.} Two pitfalls affect DeepSeek-V4-Flash on this hardware, and our numbers avoid both. First, random weights understate iteration costs by 20--35\%, because real expert weights run longer and hold the GPU at its power limit. The power limit is also why we do not overlap prefill and decode on one GPU, since both compete for the same power. Second, in vLLM~0.30 an import-order bug silently disables DeepSeek-V4's CUDA-graph breaks when the engine runs in the caller's process. We use the fixed path (\texttt{VLLM\_USE\_BREAKABLE\_CUDAGRAPH=1}).

\textbf{Release.} We will release the scheduler, the baselines, the client, all configurations and the data of every run.

\section{Detailed Results}
\label{app:details}

\begin{table}[t]
\caption{SLO goodput in req/s at $\ge$90\% attainment under the P99 TBT SLO and the worst-gap variant, both 150\,ms (Mixed; last rows: Mooncake trace); same runs as Figure~\ref{fig:main}. ``$<$$r$'': below 90\% already at the lowest rate $r$; ``$\geq$$r$'': never below 90\%; ratios against such bounds are bounds.}
\label{tab:capacity}
\centering
\footnotesize
\setlength{\tabcolsep}{2pt}
\resizebox{\columnwidth}{!}{\begin{tabular}{@{}llccccccc@{}}
\toprule
Model & TBT SLO & vLLM & Sarathi & SPF & Niyama & Medha & \sysname & $\times$best \\
\midrule
Qwen3-8B & P99 TBT & $<$0.1 & $<$0.1 & 0.10 & $<$0.1 & 0.65 & 1.54 & 2.4$\times$ \\
(GQA) & Worst gap & $<$0.1 & $<$0.1 & 0.10 & $<$0.1 & 0.18 & 1.50 & 8.4$\times$ \\
\midrule
DeepSeek-V2-Lite & P99 TBT & $<$0.125 & 0.14 & 0.19 & $<$0.125 & 0.67 & $\geq$3.00 & $\geq$4.4$\times$ \\
(MLA) & Worst gap & $<$0.125 & 0.14 & 0.19 & $<$0.125 & 0.22 & 0.48 & 2.2$\times$ \\
\midrule
DeepSeek-V4-Flash & P99 TBT & $<$0.1 & $<$0.1 & $<$0.1 & 0.13 & 0.44 & 1.29 & 2.9$\times$ \\
(CSA/HCA) & Worst gap & $<$0.1 & $<$0.1 & $<$0.1 & $<$0.1 & 0.44 & 1.29 & 2.9$\times$ \\
\midrule
Mooncake trace & P99 TBT & $<$0.15 & $<$0.15 & $<$0.15 & $<$0.15 & 0.17 & 0.45 & 2.7$\times$ \\
(V4-Flash) & Worst gap & $<$0.15 & $<$0.15 & $<$0.15 & $<$0.15 & 0.15 & 0.45 & 2.9$\times$ \\
\bottomrule
\end{tabular}
}
\end{table}

\begin{figure*}[t]
\centering
\includegraphics[width=0.8\textwidth]{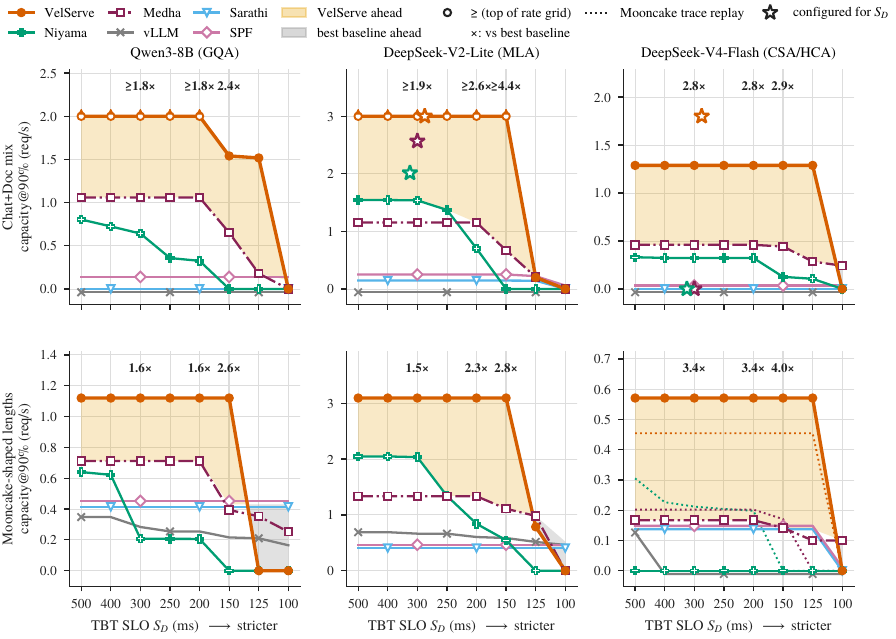}
\caption{SLO goodput as the per-request P99 TBT bound tightens, for every model and workload (TTFT SLO scale 3). Top: Mixed (labelled Chat+Doc mix). Bottom: Mooncake-shaped lengths, with the replayed trace dotted on DeepSeek-V4-Flash. Shading: \sysname ahead of the best baseline, with the ratio. Stars: configured for 300\,ms.}
\label{fig:sdgap-full-p99}
\end{figure*}

\begin{figure*}[t]
\centering
\includegraphics[width=0.66\textwidth]{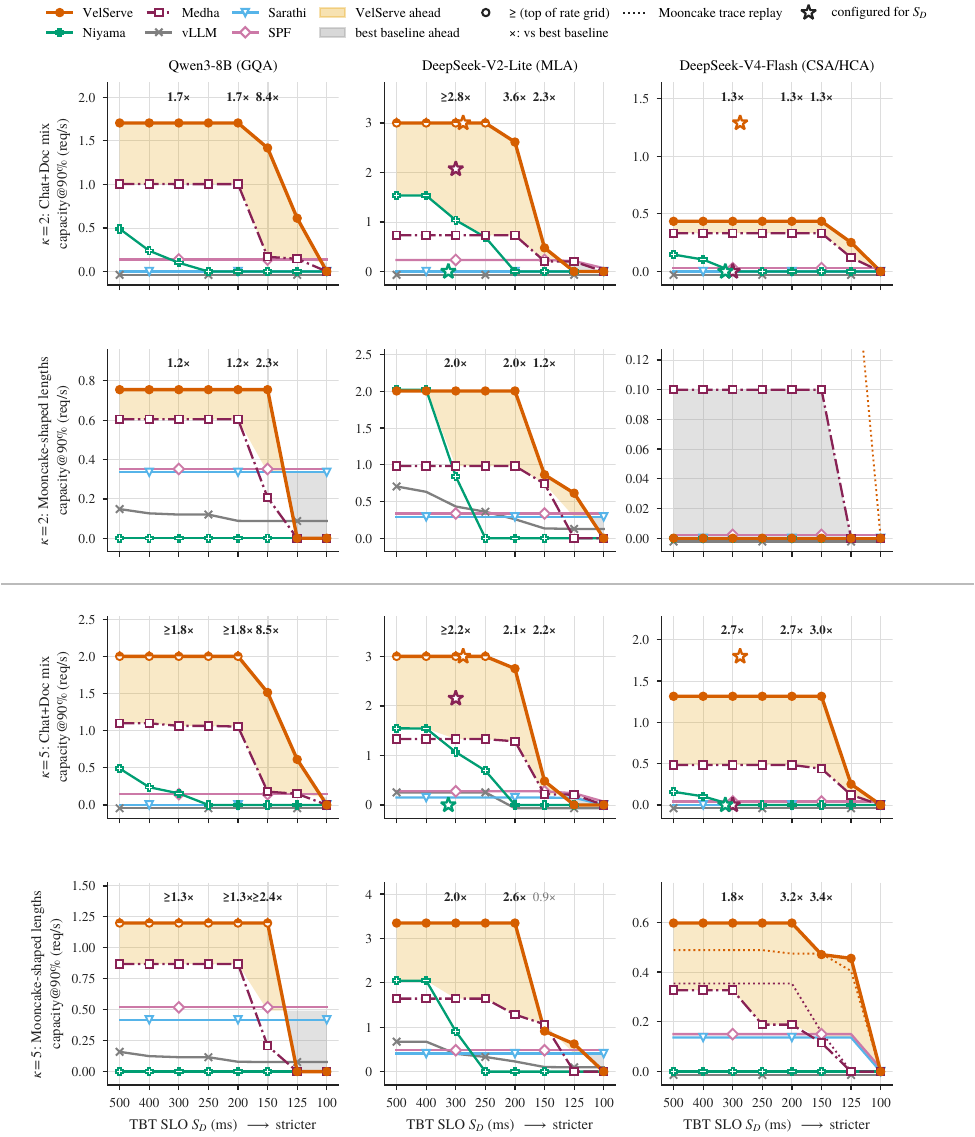}
\caption{Sensitivity to the TTFT target under the worst-gap bound. The TTFT SLO scale is 2 (top two rows) or 5 (bottom two rows). Layout as in Figure~\ref{fig:sdgap-full-p99}.}
\label{fig:kappa}
\end{figure*}

\textbf{SLO goodput.} Table~\ref{tab:capacity} lists every system's SLO goodput at 150\,ms. Medha is the strongest baseline on every model. Niyama and vLLM fall below 90\% at or near our lowest rate, and Sarathi and SPF stay low because they lose first tokens.

\textbf{TBT bound.} Figure~\ref{fig:sdgap-full-p99} sweeps the P99 bound for all six model--workload pairs, two more than Figure~\ref{fig:sdgap} shows. \sysname leads every baseline from 500 down to 150\,ms on every pair. Below 150\,ms, the bound approaches the length of host stalls themselves. \sysname still leads on three pairs at 125\,ms, and at 100\,ms it reports the SLO as infeasible (Appendix~\ref{app:design}).

\textbf{TTFT deadline.} Figure~\ref{fig:kappa} repeats the worst-gap sweep with the TTFT SLO scale tightened to 2 and relaxed to 5. A tighter scale costs \sysname most where long prompts are common, because their slack pays for its margin. In a few such cells a baseline draws level or ahead. On Mixed, \sysname's lead holds at every scale from 10 down to 2. Only at a scale of 1.2 does Niyama's deadline order serve long prompts sooner, and it then leads on DeepSeek-V4-Flash.

\begin{table}[t]
\caption{SLO goodput in req/s at $\ge$90\% attainment (Mixed, worst gap $\le150$\,ms, TTFT SLO scale 3) when the minimum TTFT deadline $S_{F0}$ is raised
from 1\,s (the SLO) to 30\,s. Same runs and notation as Table~\ref{tab:capacity}.}
\label{tab:sf0}
\centering
\footnotesize
\setlength{\tabcolsep}{3pt}
\resizebox{\columnwidth}{!}{\begin{tabular}{@{}llrrrrr@{}}
\toprule
Model & System & 1\,s & 2\,s & 5\,s & 10\,s & 30\,s \\
\midrule
Qwen3-8B & vLLM & $<$0.1 & $<$0.1 & $<$0.1 & 0.10 & 0.10 \\
 & Sarathi & $<$0.1 & 0.10 & 0.10 & 0.16 & 0.80 \\
 & SPF & 0.10 & 0.15 & 0.48 & 1.02 & 1.03 \\
 & Niyama & $<$0.1 & $<$0.1 & $<$0.1 & $<$0.1 & $<$0.1 \\
 & Medha & 0.18 & 0.18 & 0.18 & 0.18 & 0.18 \\
 & VelServe & 1.08 & 1.08 & 1.08 & 1.08 & 1.08 \\
\midrule
DeepSeek-V2-Lite & vLLM & $<$0.125 & $<$0.125 & 0.17 & 0.17 & 0.17 \\
 & Sarathi & 0.14 & 0.15 & 0.15 & 0.16 & 1.05 \\
 & SPF & 0.19 & 0.23 & 0.59 & 0.66 & 0.72 \\
 & Niyama & $<$0.125 & $<$0.125 & $<$0.125 & $<$0.125 & $<$0.125 \\
 & Medha & 0.22 & 0.22 & 0.22 & 0.22 & 0.22 \\
 & VelServe & 1.89 & 1.89 & 1.89 & 1.89 & 1.90 \\
\midrule
DeepSeek-V4-Flash & vLLM & $<$0.1 & $<$0.1 & $<$0.1 & $<$0.1 & $<$0.1 \\
 & Sarathi & $<$0.1 & $<$0.1 & 0.10 & 0.10 & 0.51 \\
 & SPF & $<$0.1 & 0.10 & 0.15 & 0.60 & 0.69 \\
 & Niyama & $<$0.1 & $<$0.1 & $<$0.1 & $<$0.1 & $<$0.1 \\
 & Medha & 0.44 & 0.44 & 0.44 & 0.44 & 0.44 \\
 & VelServe & 1.29 & 1.30 & 1.30 & 1.30 & 1.32 \\
\bottomrule
\end{tabular}
}
\end{table}

\textbf{Sarathi and SPF need very loose first-token deadlines.} They keep every TBT short, but short requests wait behind long prefills. Raising the TTFT scale does not help them, because a short prompt's deadline stays at a few seconds while a long document ahead of it prefills for tens of seconds. Table~\ref{tab:sf0} instead raises the 1-s floor of the deadline. Sarathi and SPF recover only once a short request may wait 10--30\,s for its first token.

\begin{figure}[t]
\centering
\includegraphics[width=0.86\columnwidth]{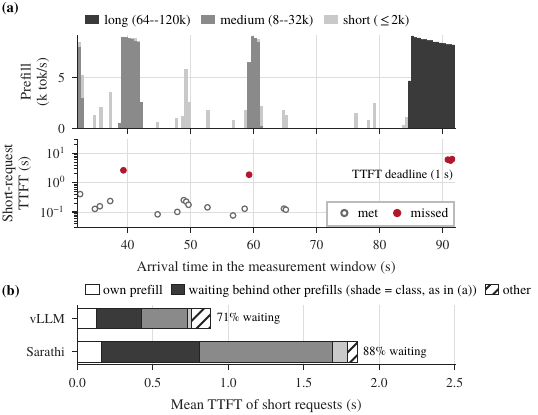}
\caption{Head-of-line blocking under FCFS (\S\ref{sec:motivation-hol}; DeepSeek-V4-Flash, Mixed). (a) Prefill throughput by prompt class, and the TTFT of each short request against its 1-s deadline. (b) Mean TTFT of short requests, split into their own prefill, waiting behind other prefills, and other delays.}
\label{fig:hol}
\end{figure}

\begin{figure*}[t]
\centering
\includegraphics[width=0.84\textwidth]{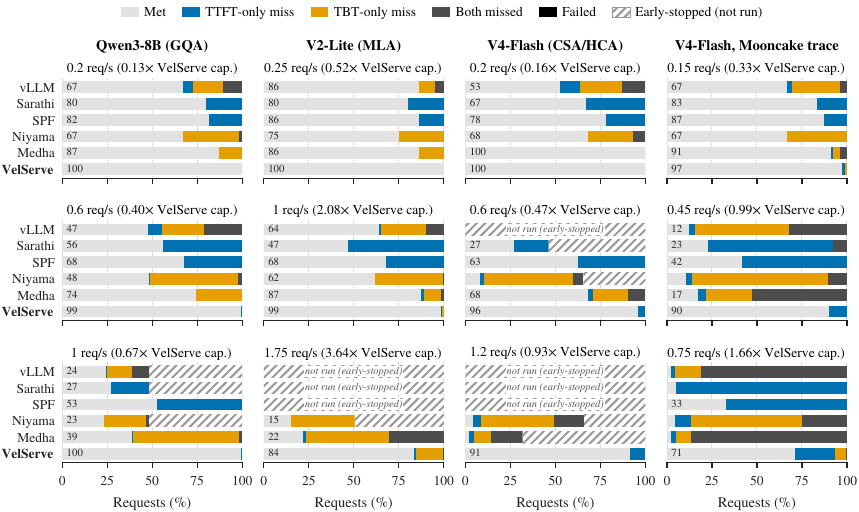}
\caption{Outcome of every request under the worst-gap SLO (150\,ms) at three loads per workload. Panel titles give the load relative to \sysname's SLO goodput (``cap.''). Numbers: \% of requests that met the SLO. Hatched: seeds stopped early below this load, counted as misses.}
\label{fig:views-miss}
\end{figure*}

\textbf{The baselines miss for the reasons of \S\ref{sec:motivation}.} Figure~\ref{fig:hol} shows head-of-line blocking under FCFS: short requests spend most of their TTFT waiting behind other prefills. Figure~\ref{fig:views-miss} splits every request by how it missed, and the two failure modes separate cleanly. Sarathi and SPF lose first tokens. Niyama and Medha lose TBT, because Niyama plans long iterations and Medha keeps no margin. vLLM loses both.

\textbf{\sysname's own misses are mostly first tokens.} Past its SLO goodput, \sysname mainly misses the first tokens of long prompts, so its attainment degrades gradually. Its few early misses on DeepSeek-V4-Flash come from iterations with no running decodes. These use vLLM's full budget, and a short prompt that arrives meanwhile can wait up to 1.5\,s.

\begin{figure*}[t]
\centering
\includegraphics[width=0.66\textwidth]{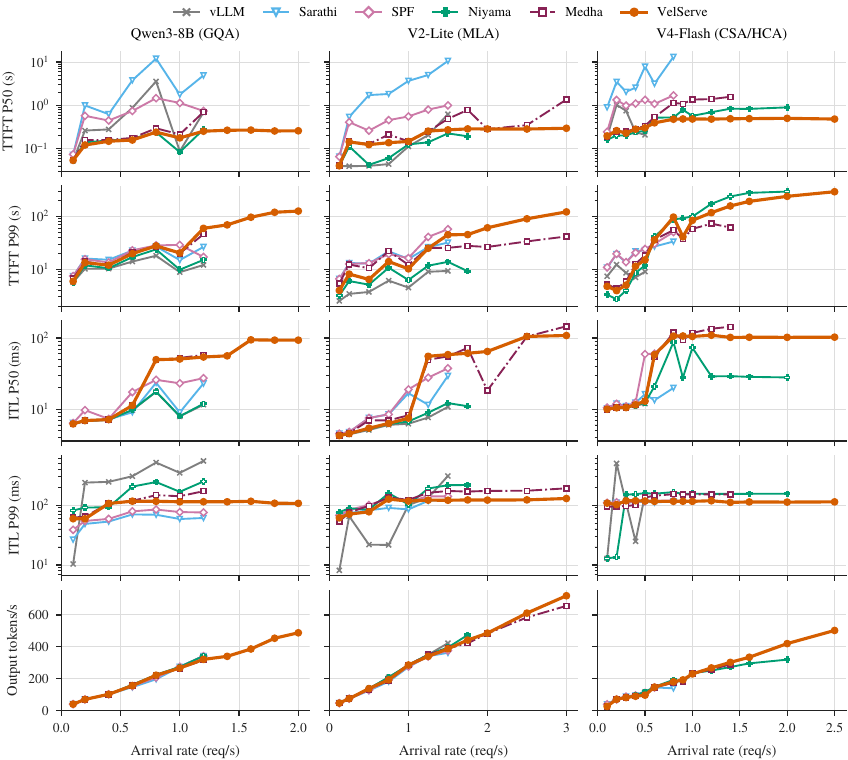}
\caption{Standard serving metrics on Mixed, with no SLO: percentiles of first-token latency (TTFT) and inter-token latency (ITL) on a log scale, and output tokens per second, against arrival rate. A line ends where the system was stopped early.}
\label{fig:views-plain}
\end{figure*}

\textbf{Standard serving metrics.} Figure~\ref{fig:views-plain} reports the same runs without any SLO. Throughput follows the offered load and is nearly the same for every system. \sysname keeps its P99 inter-token latency below the bound under load, while the baselines reach 150\,ms to over a second. Its costs show in two places. Its median inter-token latency is higher than vLLM's at moderate load, because it runs smaller iterations. Its P99 first-token latency is not lower than the baselines', because shortest-first order makes the longest prompts wait.

\textbf{Ablation (\S\ref{sec:eval-ablation}).} The arms run Qwen3-8B on Mixed at four rates with two seeds. With vLLM's 8192-token budget, many requests miss their P99 TBT target whatever the order. With \sysname's budget every request meets it, so first-come-first-served order (A2) loses only first tokens. The arms other than \sysname ran in later sessions, so vLLM's numbers differ slightly from Figure~\ref{fig:main}a.

\section{End-to-End Deadlines}
\label{app:e2e}

\begin{table}[t]
\caption{SLO goodput in req/s when the runs of Figures~\ref{fig:main} and~\ref{fig:sdgap} are scored by an end-to-end deadline: first token within the TTFT deadline, last token within it plus $(n-1)\times$ a TPOT SLO of 150\,ms. Same configurations and notation as Table~\ref{tab:capacity}.}
\label{tab:e2e}
\centering
\footnotesize
\setlength{\tabcolsep}{2pt}
\resizebox{\columnwidth}{!}{\begin{tabular}{@{}llccccccc@{}}
\toprule
Workload & Model & vLLM & Sarathi & SPF & Niyama & Medha & \sysname & $\times$best \\
\midrule
Mixed & Qwen3-8B & 0.10 & $<$0.1 & 0.10 & 0.83 & 1.06 & $\geq$2.00 & $\geq$1.8$\times$ \\
 & V2-Lite & 0.50 & 0.14 & 0.19 & 1.54 & 1.15 & $\geq$3.00 & $\geq$1.9$\times$ \\
 & V4-Flash & $<$0.1 & $<$0.1 & $<$0.1 & 0.40 & 0.46 & 1.29 & 2.8$\times$ \\
\midrule
Mooncake-sh. & Qwen3-8B & 0.44 & 0.41 & 0.43 & 0.64 & 0.71 & 1.12 & 1.6$\times$ \\
 & V2-Lite & $\geq$0.75 & 0.41 & 0.41 & 2.05 & 1.33 & 3.09 & 1.5$\times$ \\
 & V4-Flash & 0.14 & 0.14 & 0.14 & 0.50 & 0.17 & 0.57 & 1.1$\times$ \\
\bottomrule
\end{tabular}
}
\end{table}

\begin{table*}[t]
\caption{End-to-end request latency, arrival to last token, in seconds: median / P99 over completed requests (mean over seeds), and in parentheses the share of all requests (\%) that meet the end-to-end deadline of Table~\ref{tab:e2e}. Same runs as Figures~\ref{fig:main} and~\ref{fig:sdgap}; --: not run at this rate.}
\label{tab:e2e-lat}
\centering
\footnotesize
\setlength{\tabcolsep}{4pt}
\resizebox{\textwidth}{!}{\begin{tabular}{@{}lllcccccc@{}}
\toprule
Workload & Model & req/s & vLLM & Sarathi & SPF & Niyama & Medha & \sysname \\
\midrule
Mixed & Qwen3-8B & 0.4 & 3.2 / 21 (85) & 3.6 / 22 (76) & 3.3 / 22 (80) & 3.2 / 20 (99) & 3.3 / 21 (100) & 3.3 / 21 (100) \\
Mixed & Qwen3-8B & 0.8 & 12.3 / 35 (55) & 20.4 / 39 (27) & 11.3 / 42 (51) & 11.3 / 35 (97) & 13.5 / 39 (96) & 13.0 / 38 (98) \\
Mixed & V2-Lite & 0.5 & 1.9 / 10 (90) & 5.8 / 23 (71) & 4.7 / 23 (82) & 1.9 / 11 (100) & 3.1 / 20 (100) & 2.2 / 13 (100) \\
Mixed & V2-Lite & 1 & 2.5 / 12 (89) & 10.6 / 28 (47) & 8.4 / 27 (68) & 2.9 / 16 (100) & 4.8 / 22 (97) & 3.7 / 19 (100) \\
Mixed & V4-Flash & 0.2 & 6.1 / 24 (75) & 10.9 / 31 (67) & 9.2 / 30 (78) & 4.0 / 8 (93) & 4.7 / 10 (100) & 4.6 / 9 (100) \\
Mixed & V4-Flash & 0.4 & 4.7 / 15 (80) & 12.2 / 35 (62) & 8.9 / 35 (73) & 6.3 / 31 (90) & 8.5 / 35 (100) & 7.9 / 36 (95) \\
\midrule
Mooncake-sh. & Qwen3-8B & 0.4 & 3.4 / 14 (91) & 3.6 / 15 (91) & 3.7 / 15 (91) & 4.2 / 18 (97) & 4.3 / 19 (98) & 5.1 / 24 (100) \\
Mooncake-sh. & V2-Lite & 1 & -- & -- & -- & 3.8 / 17 (99) & 4.7 / 22 (97) & 4.1 / 19 (99) \\
Mooncake-sh. & V4-Flash & 0.4 & 9.1 / 43 (78) & 12.2 / 55 (64) & 12.2 / 54 (76) & 13.3 / 42 (97) & 16.6 / 48 (76) & 17.8 / 51 (94) \\
\bottomrule
\end{tabular}
}
\end{table*}

A request with $n$ output tokens meets the end-to-end deadline if its first token arrives within the TTFT deadline $D$ and its last within $D+(n-1)\SD$. This is Niyama's cumulative token deadline applied to the first and the last token. It forgives any pause for which a stream has banked time, so it is looser than a bound on each gap. Table~\ref{tab:e2e} re-scores the runs of the main evaluation this way, with every scheduler configured as in the paper.

Niyama and Medha gain the most, because most of their TBT misses disappear. Head-of-line blocking still limits vLLM, Sarathi and SPF. \sysname's misses were already first-token misses, so its SLO goodput does not drop, and it stays ahead on all six pairs. Table~\ref{tab:e2e-lat} shows what \sysname's smaller iterations cost: a longer but steady stream. Its median latency is close to Medha's and somewhat above Niyama's, which finishes sooner by running the long iterations that break the TBT bound.

\section{Iteration Overruns}
\label{app:overrun}

\begin{table}[t]
\caption{Iteration overruns on the same workload (Mooncake-shaped lengths, seed 0, $\SD=150$\,ms). Planned: \% of iterations whose predicted time exceeds $\SD$. Over: \% whose measured time exceeds $\SD$. Residual: measured minus predicted time of iterations that carry a prefill chunk, median and P99.}
\label{tab:overrun-full}
\centering\small
\resizebox{\columnwidth}{!}{%
\begin{tabular}{@{}llrrrr@{}}
\toprule
& & Planned & Over & \multicolumn{2}{c}{Residual (ms)} \\
Model & System & $>\SD$ (\%) & $\SD$ (\%) & median & p99 \\
\midrule
\csname @@input\endcsname figures/ds_overrun.tex
\bottomrule
\end{tabular}}
\end{table}

Table~\ref{tab:overrun-full} extends Table~\ref{tab:overrun} with the prediction error, or residual, of iterations that carry a prefill chunk. Niyama plans iterations longer than $\SD$ by design. Its cumulative deadlines let a stream that runs ahead bank slack, and Niyama spends that slack on iterations that last up to about a second. Medha never plans an overrun, but with no margin its prediction errors alone push a few percent of its iterations past $\SD$. \sysname's residuals stay centered near zero, and it almost never overruns. Its refit is not perfect, since rare long-context chunks remain under-predicted, but the margin absorbs them.

\section{Design Space}
\label{app:space}

\begin{figure*}[t]
\centering
\includegraphics[width=\textwidth]{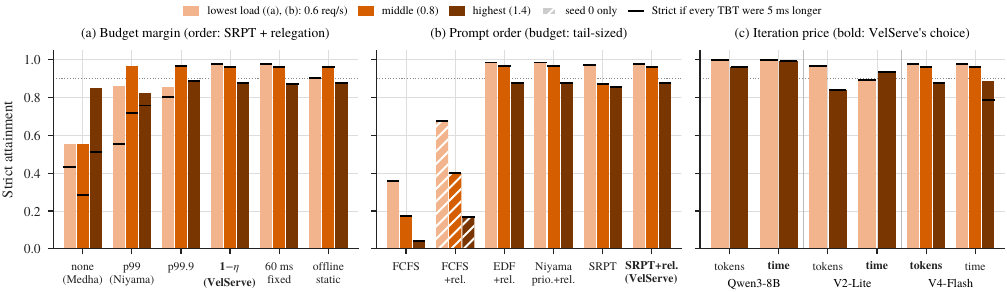}
\caption{Design space: one decision varied around \sysname, the others held at \sysname's choice (Mixed, worst-gap bound at 150\,ms, two seeds pooled). (a) Margin of the token budget (\S\ref{sec:design-budget}, DeepSeek-V4-Flash); none and p99 are Medha's and Niyama's rules applied inside \sysname. (b) Prompt order (\S\ref{sec:design-order}, DeepSeek-V4-Flash). (c) Cost estimate (\S\ref{sec:design-cost}): token counts or the context-aware time model; bold: \sysname's choice. Bars: attainment; ticks: attainment if every TBT were 5\,ms longer.}
\label{fig:space}
\end{figure*}

Figure~\ref{fig:space} replaces one of \sysname's decisions at a time with the alternatives of Table~\ref{tab:space}.

\textbf{The margin must reach past the bulk of the residuals (a).} With no margin (Medha's rule) or the 99th percentile of the residuals (Niyama's rule), attainment drops, while \sysname's margin keeps it high at every load. A fixed margin of about twice \sysname's does equally well on this model, so attainment is flat once the margin is deep enough. The difference is that \sysname finds such a margin from measurements, whereas a fixed value must be tuned per model and SLO. The ticks show how close each variant runs to the bound. Adding 5\,ms to every TBT hurts the shallow margins badly and leaves \sysname unchanged.

\textbf{Long runs expose rare stalls.} Two-minute windows see few host stalls, so we also ran 10-minute runs at 0.8\,req/s. With no margin, about two thirds of requests miss on TBT, and with the p99 margin about a third. \sysname's misses (7.6\%) all come from a single host stall that was longer than any budget could absorb, which its calibration had flagged as irreducible. This exceeds \sysname's own miss target. Its deeper margin also costs a little TTFT attainment.

\textbf{Any preemptive order that knows sizes or deadlines works (b).} With \sysname's budget, first-come-first-served order fails. Earliest-deadline-first, Niyama's priority and shortest-first all do well once late requests move to the back. The order only has to keep short prompts from waiting behind long ones. The budget is what separates \sysname from Niyama and Medha.

\textbf{The time model matters where cost grows with context (c).} On DeepSeek-V4-Flash, token counts and the time model perform the same. On DeepSeek-V2-Lite, where a token's cost grows with its context, token counts lower attainment at 0.75\,req/s from 99\% to 84\% under the P99 SLO. On Qwen3-8B, they lose little at our loads.

\textbf{A measured budget beats a hand-set one.} On DeepSeek-V4-Flash, \sysname's online budget settles slightly below the offline-calibrated static budget. With the offline budget, one seed loses requests to two slow iterations, while \sysname keeps every TBT within the bound.

\section{Measured versus Frozen Margins}
\label{app:margin}

\begin{table}[t]
\caption{SLO attainment (\%, P99 TBT SLO, Mixed, seeds 0--1) with \sysname's margin frozen, or moved into Medha. Table~\ref{tab:margin-main} shows the Qwen3-8B rows at three rates. ``--'': not run.}
\label{tab:margin}
\centering
\footnotesize
\setlength{\tabcolsep}{2.5pt}
\resizebox{\columnwidth}{!}{\begin{tabular}{@{}lrrrrrr@{}}
\toprule
Rate (req/s) & 0.6 & 0.8 & 1.0 & 1.2 & 1.4 & 1.6 \\
\midrule
\multicolumn{7}{@{}l}{\emph{Qwen3-8B}} \\
\sysname & 99.3 & 98.4 & 99.6 & 97.5 & 96.4 & 87.3 \\
\quad margin frozen offline (9.4\,ms) & 66.4 & -- & 47.6 & -- & 35.6 & -- \\
\quad margin frozen, from Mooncake-shaped lengths (29.3\,ms) & 99.3 & -- & 83.1 & -- & 91.8 & -- \\
Medha & 100.0 & -- & 57.6 & -- & 25.7 & -- \\
\quad + 29.3\,ms margin (from Mooncake-shaped lengths) & 100.0 & -- & 87.4 & -- & 51.7 & -- \\
\quad + 48\,ms margin (\sysname's value on Mixed) & 100.0 & -- & 92.2 & -- & 59.2 & -- \\
\midrule
\multicolumn{7}{@{}l}{\emph{DeepSeek-V4-Flash}} \\
\sysname & 97.8 & 96.2 & 92.6 & 91.4 & 87.9 & 85.8 \\
\quad fixed 60\,ms margin & 97.8 & 96.2 & -- & -- & 87.0 & -- \\
\quad static budget set offline & 98.5 & 96.2 & -- & -- & 87.6 & -- \\
\bottomrule
\end{tabular}
}
\end{table}

Table~\ref{tab:margin} extends Table~\ref{tab:margin-main} with more rates and with DeepSeek-V4-Flash. On DeepSeek-V4-Flash, a frozen margin and an offline static budget both match \sysname. On Qwen3-8B, a frozen margin fails on TBT, whether it is set offline or carried over from another workload, because stalls under load run deeper than the offline run saw. Moved into Medha, a carried-over margin removes only part of Medha's TBT misses, while \sysname's live value removes all of them. Medha then falls behind on first tokens, through its order. The margin is thus a part that other SLO-aware schedulers can adopt, as long as it is measured on the workload being served.

\section{Mooncake-Shaped Lengths and the Trace}
\label{app:order}

\begin{table}[t]
\caption{DeepSeek-V4-Flash with Mooncake-shaped lengths: SLO attainment (\%) pooled over three seeds.}
\label{tab:promote}
\centering
\footnotesize
\resizebox{\columnwidth}{!}{%
\begin{tabular}{@{}lrrrrrr@{}}
\toprule
Rate (req/s) & 0.4 & 0.6 & 0.8 & 1.0 & 1.4 & 1.8 \\
\midrule
\sysname & 93.7 & 89.4 & 79.8 & 76.1 & 68.0 & 60.2 \\
\sysname + promotion & 94.5 & 85.1 & 79.5 & 72.8 & 64.8 & 57.1 \\
\bottomrule
\end{tabular}}
\end{table}

\textbf{Shortest-first leaves medium and long prompts late.} With Mooncake-shaped lengths on DeepSeek-V4-Flash, nearly all of \sysname's misses are first tokens of medium and long prompts, which shortest-first order delays past their deadlines. Another order does not recover them. With Niyama's priority or earliest-deadline-first in place of shortest-first, attainment stays about the same.

\textbf{Deadline promotion does not help either.} Promotion keeps shortest-first order, but first reserves, in deadline order, the smallest share of each iteration that keeps at-risk prompts on time. It matches or slightly trails \sysname (Table~\ref{tab:promote}) and serves no more medium or long prompts on time. Up to 1.0\,req/s, every request of both versions meets its P99 TBT target, so all misses are first tokens. The margin limits how much prefill each iteration carries, and no order changes that.

\textbf{Trace.} The replayed Mooncake trace (Figure~\ref{fig:main}d) pools four slices of 150 consecutive requests, chosen before any run. One of them is the slice with the longest prompts. Each slice is time-scaled to mean rates of 0.15--0.75\,req/s. \sysname's few TBT misses on the trace come from three shared stalls, whereas most of Niyama's misses are TBT misses.

\textbf{Per-class latency.} Table~\ref{tab:classes} gives the numbers behind Figure~\ref{fig:classes}. Under \sysname, short prompts wait least. Long prompts wait about as long as under Niyama, with a similar share timing out. vLLM avoids timeouts only by making every class wait over 100\,s at the tail.

\begin{table}[t]
\caption{Per-class latency over one 40-minute window (Qwen3-8B, Mixed, 1.0\,req/s, identical arrivals). TTFT percentiles count a request with no first token within the 300-s timeout as $>$300\,s. Slowdown: TTFT divided by the prompt's prefill time on an idle engine. TBT p99: over all gaps of the class. SLO met: TTFT deadline and own P99 TBT $\le$150\,ms.}
\label{tab:classes}
\centering\small
\resizebox{\columnwidth}{!}{%
\begin{tabular}{@{}llrrrrrrr@{}}
\toprule
& & \multicolumn{2}{c}{TTFT (s)} & Slowdown & Timed & TTFT & TBT p99 & SLO \\
Class & System & p50 & p99 & p50 & out (\%) & met (\%) & (ms) & met (\%) \\
\midrule
\multirow{3}{*}{Short ($<$8k)} & vLLM & 30.6 & 107 & 706.6 & 0.0 & 19.8 & 1200 & 14.3 \\
 & Niyama & 0.3 & 1.2 & 7.1 & 0.0 & 97.0 & 454 & 11.8 \\
 & \sysname & 0.2 & 0.4 & 5.3 & 0.0 & 99.8 & 126 & 99.8 \\
\midrule
\multirow{3}{*}{Medium (8--32k)} & vLLM & 38.2 & 115 & 34.3 & 0.0 & 17.6 & 1216 & 10.5 \\
 & Niyama & 1.8 & 214 & 1.7 & 0.4 & 90.5 & 459 & 8.8 \\
 & \sysname & 1.7 & $>$300 & 1.6 & 2.9 & 91.0 & 121 & 91.0 \\
\midrule
\multirow{3}{*}{Long (64k+)} & vLLM & 39.9 & 111 & 3.7 & 0.0 & 44.5 & 1202 & 7.3 \\
 & Niyama & 26.6 & $>$300 & 2.6 & 20.0 & 61.8 & 466 & 9.1 \\
 & \sysname & 23.3 & $>$300 & 2.4 & 19.1 & 66.4 & 127 & 66.4 \\
\midrule
\multirow{3}{*}{All} & vLLM & 33.1 & 110 & 288.8 & 0.0 & 20.5 & 1202 & 13.3 \\
 & Niyama & 0.4 & 263 & 5.9 & 1.0 & 94.2 & 456 & 11.1 \\
 & \sysname & 0.3 & $>$300 & 4.6 & 1.4 & 96.6 & 125 & 96.6 \\
\bottomrule
\end{tabular}}
\end{table}

\section{Multi-Turn Serving and the Agent Trace}
\label{app:multiturn}

\begin{figure*}[t]
\centering
\includegraphics[width=\textwidth]{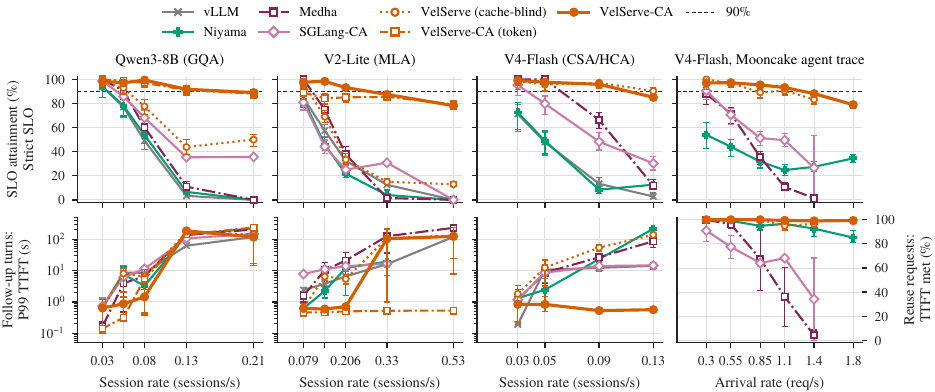}
\caption{Prefix caching. Left three columns: Multi-turn sessions over a long document, mean of two seeds. Right: DeepSeek-V4-Flash replaying the Mooncake agent trace. Top: attainment under the worst-gap bound (150\,ms), with 95\% Wilson intervals. Bottom: P99 TTFT of follow-up turns (log scale); right: share of prefix-reusing requests that meet their TTFT deadline.}
\label{fig:e5}
\end{figure*}

\begin{table}[t]
\caption{SLO goodput with prefix caching at $\ge$90\% attainment (Multi-turn: sessions/s; agent trace, last two rows: req/s), and cache-aware \sysname's ratio over the strongest baseline (vLLM, Niyama, Medha, SGLang-CA = SGL-CA). Strict: the worst-gap bound of 150\,ms used in the text; Relaxed: a looser TBT bound. n/a: not run.}
\label{tab:e5}
\centering
\footnotesize
\setlength{\tabcolsep}{2pt}
\resizebox{\columnwidth}{!}{\begin{tabular}{@{}llcccccccc@{}}
\toprule
 & & \multicolumn{4}{c}{Baselines} & \multicolumn{3}{c}{\sysname} & \\
\cmidrule(lr){3-6}\cmidrule(lr){7-9}
Model & SLO & vLLM & Niyama & Medha & SGL-CA & blind & CA & CA-tok. & $\times$best \\
\midrule
Qwen3-8B & Strict & 0.036 & 0.036 & 0.055 & 0.047 & 0.060 & 0.184 & 0.161 & 3.3$\times$ \\
(GQA) & Relaxed & 0.036 & 0.037 & 0.084 & 0.047 & 0.093 & 0.184 & 0.161 & 2.2$\times$ \\
\midrule
V2-Lite & Strict & 0.051 & $<$0.048 & 0.105 & 0.053 & 0.093 & 0.276 & 0.070 & 2.6$\times$ \\
(MLA) & Relaxed & 0.052 & 0.051 & 0.111 & 0.053 & 0.150 & 0.382 & 0.070 & 3.4$\times$ \\
\midrule
V4-Flash & Strict & $<$0.03 & $<$0.03 & 0.062 & 0.037 & $\geq$0.130 & 0.112 & n/a & 1.8$\times$ \\
(CSA/HCA) & Relaxed & $<$0.03 & 0.031 & 0.079 & 0.037 & $\geq$0.130 & $\geq$0.130 & n/a & $\geq$1.6$\times$ \\
\midrule
V4-Flash & Strict & n/a & $<$0.3 & $<$0.3 & 0.307 & 0.821 & 1.295 & n/a & 4.2$\times$ \\
(agent) & Relaxed & n/a & $<$0.3 & 0.524 & 0.307 & 1.182 & 1.295 & n/a & 2.5$\times$ \\
\bottomrule
\end{tabular}
}
\end{table}

This section uses the worst-gap bound. With prefix caching, a follow-up turn in a long-document session recomputes only its new tokens. A cache-blind size ranks it by its full prompt, so it waits behind fresh documents. We run the Multi-turn workload with prefix caching on for every system. We add SGLang-CA, SGLang's shortest-prefill-first by uncached tokens~\cite{sglang2026spf}, ported to the same vLLM build. All systems reuse a similar share of tokens from the cache. They differ in \emph{when} they schedule the uncached work (Figure~\ref{fig:e5}).

\textbf{Cache-aware sizes keep follow-up turns fast.} Until well into overload, cache-aware \sysname keeps the follow-up turns' P99 TTFT within about 1.5\,s on every model. With cache-blind sizes, these turns wait behind fresh documents, and under the baselines they wait up to minutes. Cache-aware \sysname's SLO goodput is 1.8--3.3$\times$ that of the strongest baseline, Medha (Table~\ref{tab:e5}). Niyama again loses on TBT, through the long iterations that its banked slack allows. On MLA, the size must be estimated in time, and forced token sizes lower attainment.

\textbf{The agent trace confirms the effect.} The Multi-turn sessions are synthetic, so we also replay the Mooncake tool-agent trace on DeepSeek-V4-Flash, whose requests share prefixes as recorded (Figure~\ref{fig:e5}, right). Cache-aware \sysname sustains 4.2$\times$ the SLO goodput of SGLang-CA, and Niyama and Medha are below 90\% at our lowest rate. The shared prefixes here are mostly short system and tool prompts, so cache-aware sizes matter less than for long documents.

\section{Disaggregated Serving}
\label{app:disagg}

\begin{table*}[t]
\caption{Disaggregated serving of DeepSeek-V4-Flash (\S\ref{sec:eval-disagg}). Attainment: share of requests that meet the SLO, the P99 TBT target ($\le$150\,ms) and the TTFT deadline. Latency percentiles are over requests; TPOT is a request's mean time per output token after the first. Gaps: share of all token gaps longer than $\SD$. Italic rows give the change relative to the row named, $100(x/x_{\mathrm{ref}}-1)$: higher is better for attainment, lower for latency and gaps. The ablation ran in a separate window, so compare its rows only with each other.}
\label{tab:pd}
\centering
\footnotesize
\setlength{\tabcolsep}{3pt}
\begin{tabular}{@{}l rrr rrr rrr rrr r@{}}
\toprule
 & \multicolumn{3}{c}{Attainment (\%)} & \multicolumn{3}{c}{TTFT (s)} & \multicolumn{3}{c}{TTLT (s)} & \multicolumn{3}{c}{TPOT (ms)} & Gaps \\
\cmidrule(lr){2-4}\cmidrule(lr){5-7}\cmidrule(lr){8-10}\cmidrule(lr){11-13}
System & SLO & TBT & TTFT & P50 & P90 & P99 & P50 & P90 & P99 & P50 & P90 & P99 & $>\SD$ (\%) \\
\midrule
\multicolumn{14}{@{}l}{\emph{20 minutes, 1,200 requests}} \\
vLLM & 15.6 & 19.6 & 74.6 & 2.38 & 7.35 & 17.57 & 14.6 & 25.5 & 44.4 & 33.7 & 40.7 & 60.9 & 3.25 \\
Niyama & 14.8 & 18.6 & 78.9 & 2.30 & 5.62 & 18.57 & 14.6 & 25.5 & 43.7 & 34.2 & 41.5 & 64.4 & 3.36 \\
\sysname & 35.1 & 45.3 & 75.1 & 3.25 & 7.10 & 19.83 & 14.2 & 24.2 & 42.4 & 30.4 & 36.1 & 40.5 & 1.57 \\
\emph{\quad vs.\ vLLM} & \emph{+125\%} & \emph{+131\%} & \emph{+1\%} & \emph{+37\%} & \emph{$-$3\%} & \emph{+13\%} & \emph{$-$2\%} & \emph{$-$5\%} & \emph{$-$5\%} & \emph{$-$10\%} & \emph{$-$11\%} & \emph{$-$34\%} & \emph{$-$52\%} \\
\emph{\quad vs.\ Niyama} & \emph{+137\%} & \emph{+144\%} & \emph{$-$5\%} & \emph{+41\%} & \emph{+26\%} & \emph{+7\%} & \emph{$-$2\%} & \emph{$-$5\%} & \emph{$-$3\%} & \emph{$-$11\%} & \emph{$-$13\%} & \emph{$-$37\%} & \emph{$-$53\%} \\
\midrule
\multicolumn{14}{@{}l}{\emph{Gate ablation, 300 requests on warm \sysname servers}} \\
gate off & 14.7 & 20.0 & 75.3 & 2.57 & 5.84 & 21.55 & 14.4 & 24.5 & 44.0 & 32.5 & 39.8 & 51.0 & 3.15 \\
gate on & 44.7 & 59.7 & 71.7 & 3.64 & 7.34 & 18.44 & 14.1 & 24.0 & 45.1 & 28.7 & 36.0 & 38.4 & 1.33 \\
\emph{\quad on vs.\ off} & \emph{+205\%} & \emph{+198\%} & \emph{$-$5\%} & \emph{+42\%} & \emph{+26\%} & \emph{$-$14\%} & \emph{$-$2\%} & \emph{$-$2\%} & \emph{+3\%} & \emph{$-$12\%} & \emph{$-$10\%} & \emph{$-$25\%} & \emph{$-$58\%} \\
\bottomrule
\end{tabular}

\end{table*}

\begin{table}[t]
\caption{Parameters of the disaggregated extension and how each is set.}
\label{tab:pd-params}
\centering\footnotesize
\setlength{\tabcolsep}{3pt}
\begin{tabular}{@{}p{2.15cm}p{2.0cm}p{3.6cm}@{}}
\toprule
Parameter & Value here & Set by \\
\midrule
Epoch period $T$ & 4\,s & median of $n\,t_d/\max(k(n),1)$ (2.15\,s), rounded up on a $0.5\cdot2^j$\,s grid \\
Cap on $T$ & 4\,s & hand; did not bind \\
Gate off & join $\le\SD-2$\,ms & join cost measured online (181--183\,ms) \\
Opening window & 0.15\,s & fixed \\
Urgent sub-epoch & 0.5\,s & fixed \\
Urgent slack & 0.25\,s (rescue: 0.05\,s) & fixed \\
Prefill step & 8 sequences, 6,144 tokens & hand; same for every scheduler \\
\bottomrule
\end{tabular}
\end{table}

\textbf{Setup.} One node has eight RTX PRO 6000 GPUs connected over PCIe. Prefill and decode each run on four of them, as data-parallel ranks with expert parallelism, following vLLM's single-node P/D recipe for DeepSeek-V4-Flash. KV caches move over NIXL, and a router assigns requests round-robin. Every scheduler gets the same prefill step limit (Table~\ref{tab:pd-params}). Each system starts on fresh servers, warms up on other rows of the trace, and then serves the same 1,200 consecutive Mooncake requests over 20 minutes, at a mean of 1\,req/s. Niyama is our version (Appendix~\ref{app:niyama}), which attained slightly more than a port of the authors' code in an offline replay.

\textbf{Why a join stalls every rank.} A joining request arrives with its prompt's KV cache, but its last prompt token must still be processed. The join step therefore mixes a prompt token with the decodes, and it cannot replay the decode-only CUDA graph. It launches about 60 times as many kernels, so the host's launch rate sets its length, whatever the batch size and however many requests join. Expert parallelism then spreads the stall. The ranks agree on one execution mode per step so that their collectives match, and the collectives in every MoE layer hold each rank to the slowest one. A join on one rank therefore slows all four. The gate does not depend on this profile, because it measures the join cost online.

\textbf{The epoch period follows from the SLO.} Each opening of the gate stalls every stream once. A request with $n$ output tokens may have $k(n)$ late gaps under its P99 target, and it decodes for about $n\,t_d$ seconds. It therefore meets its target if it sees at most one opening every $n\,t_d/k(n)$ seconds. \sysname sets the epoch period $T$ to the median of this value over recent requests, so no output length is predicted. It rounds the median up on a grid of powers of two, which keeps ranks aligned even when their estimates differ. Rounding up errs toward fewer stalls, at the cost of a longer wait for the first token. The gate stays off whenever a join fits within $\SD$.

\textbf{Urgent openings protect first tokens.} Holding a request spends its TTFT slack, so each request carries its TTFT deadline from the prefill instance to the decode ranks. When a rank's most urgent ready request cannot wait for the next epoch, the rank opens early. Every rank with ready requests joins that opening, since all ranks pay the stall anyway. About half of all stalls come from such openings, which trade TBT for first tokens.

\textbf{The prefill instance keeps shortest-first order.} No decode shares the prefill GPUs, so there is no TBT to protect there and no reason to shrink prefill steps. The prefill instance serves the shortest prompt first and moves requests that will miss anyway to the back, as in colocated serving. The same plugin runs on both instances.

\textbf{The gate explains the gain.} After the main run, we replayed the first 300 requests twice on the same servers, with the gate off and then on. With the gate off, \sysname performs like vLLM and Niyama on the same requests, so its prefill order contributes little (Table~\ref{tab:pd}, ablation rows). This is a single pair of windows, so it gives no run-to-run variance.

\textbf{Sensitivity.} \sysname's lead holds while the TBT bound is shorter than a join, up to about 165\,ms. Once the bound exceeds every join, attainment equals TTFT attainment, and the gate should be off. \sysname also leads under the worst-gap bound and at TTFT SLO scales from 2 to 4, and trails vLLM slightly at 1.5. Answers of every length gain, and long answers see about half as many stalls as under the baselines. The TTFT cost falls on short prompts, which meet their deadline somewhat less often.

\textbf{Scope.} These results come from one node, one split of four prefill and four decode GPUs, one model, one engine and one trace.

\section{Baselines}
\label{app:baselines}

All systems run on the same vLLM build and flags, and all capture CUDA graphs for iterations of up to 2304 tokens. vLLM's default captures only up to 512 tokens, which would leave larger iterations without a graph and distort any comparison of budgets. For Sarathi-Serve and Niyama, we also ran the authors' code. It attained less than our versions under our SLO (Appendix~\ref{app:official}), so the paper compares against ours.

\subsection{Sarathi-Serve}
\label{app:sarathi}

Sarathi-Serve~\cite{agrawal2023sarathi,agrawal2024sarathiserve} fills each iteration with every running decode plus prefill chunks, oldest request first, up to a fixed token budget profiled offline. vLLM's chunked prefill already works this way, so our Sarathi is vLLM with a static budget sized from our cost measurements. We pick the largest budget that keeps a full iteration at the longest context within 150\,ms, including the measured tail of host delays. This gives 640 tokens on DeepSeek-V4-Flash and 336 on Qwen3-8B. On DeepSeek-V2-Lite, no budget meets 150\,ms at 123k-token contexts, so it gets the 200-ms budget of 592 tokens. Sarathi-Serve's own profiling rule gives a larger budget on Qwen3-8B, which attains about the same. Larger budgets do not rescue Sarathi either, because it serves prompts oldest first.

\textbf{Authors' code.} We ran the official repository on Qwen3-8B with the same budget and target. We added Qwen3 through its Llama code path and upgraded FlashInfer, whose pinned version no longer installs. It fails the same way as our version, on first tokens with no TBT over 150\,ms. Its engine is slower per iteration, though, so more first tokens are late (Table~\ref{tab:official}). The policy is reproduced, and the paper runs it on the faster engine.

\subsection{Niyama}
\label{app:niyama}

Niyama~\cite{goel2025niyama} schedules to deadlines instead of a fixed budget. It sizes each iteration's prefill to the slack of the most urgent running decode, and orders prompts by a mix of earliest deadline and least remaining work. It moves requests that will miss anyway to the back, and preempts prefills at chunk boundaries. Its decode deadlines are cumulative: token $n$ is due at the TTFT plus $n$ TBTs. This sets a token rate rather than bounding each gap.

Our version is a vLLM plugin that keeps all four parts. It differs from the paper as follows.
\begin{itemize}
\setlength\itemsep{0pt}
\item It splits each iteration's prefill budget across several prompts instead of one, which leaves less of the budget unused.
\item It predicts iteration time with a context-aware linear model fit on the same hardware, instead of a random forest trained on simulator profiles.
\item It subtracts a margin, the 99th percentile of its prediction error measured on static-budget runs. The paper states no margin, and Table~\ref{tab:space} marks it as our addition.
\item It moves a request to the back as soon as its deadline becomes unreachable, and it does not check whether a preempted prefill can still meet its own deadline.
\end{itemize}
A small sweep of the priority weight showed no effect, so we use 1.

\textbf{Authors' code.} We ran Niyama's release on Qwen3-8B with our SLO in place of theirs and every request in its interactive class. We also re-fitted its latency predictor on this engine with the release's own tool. The release misses mainly on TBT, with worst gaps of hundreds of milliseconds, and the re-fitted predictor changes nothing. The cause is the cumulative deadlines. Its decodes run far ahead of one token per 150\,ms and bank slack. When a chunk is sized, the most urgent decode usually has far more than $\SD$ of slack, so Niyama plans iterations past $\SD$ by design, and no predictor can fix this. Our version keeps the same deadlines and attains more than the release (Table~\ref{tab:official}).

\subsection{Medha}
\label{app:medha}

Medha~\cite{agrawal2024medha} orders prefills by least relative slack, the time left before a request's deadline divided by its remaining work, and preempts them at chunk boundaries. Each iteration takes every decode and adds prefill chunks while the predicted iteration time stays below a target. Long prefills leave part of the target to short ones. We implemented its single-replica scheduler ourselves and set the choices that the paper leaves open as follows.
\begin{itemize}
\setlength\itemsep{0pt}
\item The iteration-time target is $\SD$, with no margin, and the deadline is our TTFT deadline.
\item A prompt is long above 8k tokens, and a long prefill gives up at most 40\% of the iteration, following the paper's examples.
\item Only long prefills give up part of the iteration. Applied to every prefill, the rule would cap nearly every iteration at 60\% of the target under our deadlines.
\item The predictor is our most accurate offline cost model, which errs by a few milliseconds at the 150-ms target.
\end{itemize}
We enforce each planned chunk exactly through vLLM's per-request chunk limit.

\subsection{Comparison with the Official Artifacts}
\label{app:official}

\begin{table}[!htb]
\caption{Official artifacts versus our versions: Qwen3-8B, Mixed, seed 0. Attainment under the worst-gap bound at 150/200/300\,ms; misses at 150\,ms split into TTFT only, TBT only and both.}
\label{tab:official}
\centering
\footnotesize
\setlength{\tabcolsep}{3pt}
\resizebox{\columnwidth}{!}{\begin{tabular}{@{}llccccc@{}}
\toprule
req/s & System & 150 & 200 & 300 & Misses &  worst TBT p99 (ms)\\
\midrule
0.2 & Sarathi-Serve, official & 0.96 & 0.96 & 0.96 & 1/0/0 & 40\\
    & Sarathi, ours & 1.00 & 1.00 & 1.00 & 0/0/0 & 31\\
    & Niyama, official (tuned) & 0.69 & 0.88 & 1.00 & 0/8/0 & 273\\
    & Niyama, ours & 0.81 & 0.81 & 0.85 & 0/5/0 & 494\\
\midrule
0.6 & Sarathi-Serve, official & 0.56 & 0.56 & 0.56 & 27/0/0 & 144\\
    & Sarathi, ours & 0.77 & 0.77 & 0.77 & 14/0/0 & 85\\
    & Niyama, official (tuned) & 0.35 & 0.39 & 0.73 & 0/40/0 & 652\\
    & Niyama, ours & 0.61 & 0.68 & 0.73 & 1/20/3 & 739\\
\bottomrule
\end{tabular}}
\end{table}

Table~\ref{tab:official} compares the authors' Sarathi-Serve and Niyama with our versions on Qwen3-8B under the worst-gap bound. Our versions attain more at both loads. This is a limited check, with one seed and no iteration-by-iteration comparison of scheduling decisions.

\end{document}